\documentclass[journal,12pt]{elsarticle}
\usepackage{graphicx} 
\usepackage{float}
\usepackage{tikz}
\usepackage{amssymb}
\usepackage{bm}
\usepackage{booktabs}
\usepackage{amsmath,epsfig}
\usepackage{array}
\usepackage{gensymb}
\usepackage{eurosym}
\usepackage{url}
\usepackage{enumitem}
\usepackage[linesnumbered,ruled,vlined]{algorithm2e}
\usepackage{multirow}
\usepackage{makecell}
\usepackage{lineno}
\usepackage{colortbl}
\usepackage{caption}
\usepackage{xurl}

\begin{document}

\begin{frontmatter}

\title{A Machine Learning Approach for the Efficient Estimation of Ground-Level Air Temperature\\in Urban Areas}

\author[address1]{Iñigo Delgado-Enales}
\author[address1]{Joshua Lizundia-Loiola}
\author[address1]{\\Patricia Molina-Costa}
\author[address1,address3]{Javier~Del~Ser\corref{mycorrespondingauthor}}
\cortext[mycorrespondingauthor]{Corresponding author. Parque Tecnologico de Bizkaia, 700, 48160 Derio, Bizkaia, Spain. Phone: +34946430850.}
\ead{javier.delser@tecnalia.com}

\address[address1]{TECNALIA, Basque Research and Technology Alliance (BRTA), 48160 Derio, Spain}
\address[address3]{University of the Basque Country (UPV/EHU), 48013 Bilbao, Spain}

\begin{abstract}
The increasingly populated cities of the 21st Century face the challenge of being sustainable and resilient spaces for their inhabitants. However, climate change, among other problems, makes these objectives difficult to achieve. The Urban Heat Island (UHI) phenomenon that occurs in cities, increasing their thermal stress, is one of the stumbling blocks to achieve a more sustainable city. The ability to estimate temperatures with a high degree of accuracy allows for the identification of the highest priority areas in cities where urban improvements need to be made to reduce thermal discomfort. In this work we explore the usefulness of image-to-image deep neural networks (DNNs) for correlating spatial and meteorological variables of a urban area with street-level air temperature. The air temperature at street-level is estimated both spatially and temporally for a specific use case, and compared with existing, well-established numerical models. Based on the obtained results, deep neural networks are confirmed to be faster and less computationally expensive alternative for ground-level air temperature compared to numerical models.


\end{abstract}

\begin{keyword}

Urban Heat Island \sep Street-level Temperature \sep Data Modeling \sep Deep Neural Networks

\end{keyword}

\end{frontmatter}


\section{Introduction}\label{sec:intro}

The concept of sustainable cities, coined at the end of the 20th century, has been evolving until today, where it has become more relevant to the point that the UN has introduced it as one of The Sustainable Development Goals for 2030 \cite{un2023report}. Indicator 11 of the UN Sustainable Development Goals \cite{un2023report} defines sustainable cities as those that are resilient, safe and sustainable. This sustainability is measured in different characteristics that the city possesses such as the urban fabric, land use, public transport or environmental factors \cite{sustainablecity}. Achieving this goal is of major relevance due to the rise of cities, provoked by the migration trend from rural to urban areas that has taken place in recent decades, especially in Africa and Asia \cite{wu1996rural,lall2006rural}. According to the United Nations, this tendency will continue and by 2050, 70\% of the world's population will live in urban areas \cite{un2018}. However, cities faces several challenges in their way to be more sustainable and livable spaces. A major obstacle is the pronounced acceleration in global warming, with significant implications for the Earth’s climate system to date. This shift is evident in the increase in global average temperature, alongside a rise in the frequency of annual heatwave days \cite{LancetReport2020}. 

The consequences of the hotter urban climate can be suffered in several ways. A major problem directly affecting sustainability lies in the rising energy consumption of housing and industry. As heat consumption is closely related to temperature variations influenced by climate change, this poses a pressing concern for sustainable practices. For example, 40\% of all energy consumed in Europe is consumed by the housing stock, as well as accounting for 36\% of all CO\textsubscript{2} emissions \cite{heatclimate}. In addition, higher temperatures in cities prevent particulate pollutants from being released into the atmosphere, worsening air quality problems in urban environments \cite{pollutionsus, urbanization_pollution_suhi}. Moreover, this phenomena also affects sorely to the individuals. People facing high temperatures and heat waves can experience various adverse health effects. These consequences encompass increased susceptibility to heat stress and heatstroke, leading to illness and death, along with deterioration in respiratory and cardiovascular health \cite{health1,health2}. Over the past two decades, the mortality rate associated with heat-related incidents has increased markedly by 53.7\% for people aged 65 years and older \cite{LancetReport2020}. Another clear consequences of high temperatures is the deficit on productivity. The economic impact of heatwaves is the loss of 302 billion working hours in 2019, representing a rising of 51.8\% compared to the year 2000 \cite{LancetReport2020}. 

This rise in urban temperatures is a pressing issue, primarily due to the UHI phenomenon \cite{uhi}. The UHI is an anomaly in which the air temperature (T\textsubscript{a}), measured at a height of 2 meters above street level, is higher in urban areas compared to surrounding rural regions, with temperature increases reaching up to 7\degree C \cite{Urbclim}. This elevation in T\textsubscript{a} is driven by several factors, predominantly human activities such as traffic, industrial operations, and the use of air conditioning, which contribute to anthropogenic heat generation \cite{uhi,anthropogenicgreenhouse,oke1982}. In addition to the general UHI effect, urban hotspots represent localized areas within a city that experience even higher thermal gradients than their immediate surroundings, influenced by micro-scale factors such as building density, surface materials, and the scarcity of vegetation. This phenomenon emphasizes the complex interaction between urbanization and localized thermal dynamics. Furthermore, the increase in air temperature is closely related to land use patterns and their transformation by human activities, such as the paving of green spaces, the development of industrial zones, and the construction of new buildings. The absence of green spaces and water bodies, which are essential for natural cooling processes, further exacerbates the rise in temperatures \cite{urbanization_pollution_suhi,landcoverandUHI,karimi2023new}. Urban planning plays a critical role in determining the extent of UHI effects and hotspots. Studies have demonstrated that a city's morphology, such as the orientation and configuration of street canyons, affects heat mitigation. Poorly oriented or obstructed street canyons can impede air flow and prevent effective ventilation, thereby failing to reduce air temperatures \cite{steeneveld2011quantifying,zhou2018impact}.

All these factors underscore the need for strategic interventions to mitigate the effects of UHI and hotspots, a growing concern in resilient and sustainable cities. Eco-friendly urban planning and urban regeneration initiatives are among the most effective measures to address this issue, contributing to the broader goals of social equity, economic viability, and environmental protection. In this context, having precise data on T\textsubscript{a} values at the street level would enable urban planners to identify areas with higher heat stress, providing essential insights for urban regeneration and mobility planning, and further aligning with the objectives of fostering resilient, sustainable urban environments.

Historically, methods applied to predict T\textsubscript{a} have predominantly relied on computationally intensive numerical models. However, nowadays with new technologies arising in the urban domain such as the urban digital twins, there is a need for fast computational speed for a near-real time response of decision makers. Moreover, having short time periods of data processing is very relevant in urban areas where fast response and planning capacity is needed. Therefore, trained AI models can provide the necessary speed with current computational resources, fulfilling this gap. 

However, building a generalizable AI model that estimates the T\textsubscript{a} in different cities is not straightforward. In this paper we will discuss the potential of an AI modeling approach for spatio-temporal data to estimate temperature from simulations of an urban model and under controlled meteorological conditions. The proposed methodology addresses the estimation of T\textsubscript{a} with a spatial resolution of 100 meters using a U-Net architecture. The model has been tested in the metropolitan area of Bilbao, Spain. The results have been compared both with the reference numerical model and with data collected from meteorological stations, aiming to provide evidence to the responses of the three research questions (RQs) posed in what follows:
 \begin{itemize}[leftmargin=*]
    \item \textbf{RQ1:} Can a data-based model achieve accurate T\textsubscript{a} estimations that are close to those produced by a numerical model? 
    \item \textbf{RQ2:} Can a data-based model consistently estimate temperature over time?
    \item \textbf{RQ3:} Can a data-based model estimate temperature with a fine-grained resolution so that hotspots in cities can be identified? 
\end{itemize}

The subsequent sections of this paper are organized as follows: Section \ref{sec:background} provides a concise review of relevant literature related to the research. Section \ref{sec:framework} presents the different datasets used and explains the methodology behind the research. Section \ref{sec:results} answers the three RQs postulated above. Finally, Section \ref{sec:Conclusions} summarizes the concluding remarks drawn from the results and the future research directions. 

\section{Background}\label{sec:background}

This background section provides the necessary context for the research, focusing on two important topics. Subsection \ref{unet} explores the principles and advancements in image segmentation, with a particular emphasis on the U-Net architecture, and its applicability to solve urban-related problems. The second one, Subsection \ref{Ta review}, examines the different approaches towards estimating ground-level air temperature patterns, including current modeling approaches. Additionally, we provide a formal statement of its contributions to the state-of-the-art, highlighting how the research builds upon and advances existing knowledge in these areas.

\subsection{Image Segmentation Task: U-Net Architecture}\label{unet}

 The encoder-decoder architecture represents a fundamental framework for tensor-to-tensor modeling (as in semantic segmentation or image-to-image prediction tasks). It is based on a concatenation of convolutional networks, divided into two sections. On the one hand, there is the encoder which is in charge of reducing the dimensionality of the input sequence and extracting high quality characteristics. On the other hand, there is the decoder, which gradually reconstructs the shape until the target sequence is obtained, either a reconstruction of the image at the input of the encoder, predicted pixelwise categories or any other measurement of interest. Although this technology is mainly used for image or semantic segmentation task, examples of encoder-decoders for other uses can also be found in text summarization \cite{encodecosummary,encoderdecosummary2,encodersummary} or machine translation \cite{machinetranslation1,machinetranslation2}, to mention a few exemplary applications. 

Focusing on image prediction and segmentation tasks, diverse architectures can be devised. One of the pioneering architectures for semantic segmentation is the Fully Convolutional Network (FCN) \cite{fcn} which replaces the fully connected layers with convolutional layers to be able to predict outputs from different arbitrary-sized images/inputs. Another well known architecture is SegNet  \cite{segnet}, where the substantial difference with other models is that the decoder takes advantage of the indices calculated in the max pooling operation of the encoder to reconstruct the image in a non-linear way. This method significantly reduces both the computational cost and the computational time. Another prominent frameworks are DeepLab \cite{chen2017deeplab}, Pyramid Scene Parsing Network (PSPNet) \cite{zhao2017pyramid} and U-Net \cite{2015unet}. Moving forward, the attention will be directed towards the latter architecture, which is pertinent to the research line of this paper.

The U-Net architecture traces back to 2015 when Ronneberger et al. \cite{2015unet} proposed the use of convolutional neural networks (CNNs) for biomedical image segmentation tasks. Thereafter, it has become a very popular architecture in the area of image segmentation, especially in problems subject to data scarcity. Since its inception, U-Net models have been widespread in the biomedical domain and used for medical applications, largely due to their effectiveness in accurately delimiting objects. Its major contributions to the field are tumour detection \cite{unettumour,tumourunet2,tumourunet3} and organ segmentation in scanners \cite{medicine2020denseunet,medicine2022swinUnet,medicine2022afterunet}. Despite its success in the medical domain, the state-of-the-art of U-Net extends beyond biomedical applications. In recent years, due to its pixel-wise segmentation accuracy, researchers have increasingly applied U-Net architectures to address other types of problems. In infrastructure maintenance, it has been used to detect road pavement defects  \cite{unet_pavement} and in the same way, in textile manufacturing, for fabric defect detection \cite{unet_fabric}. Furthermore, its adoption for the analysis of satellite imagery improves the accuracy of spatial information retrieval like in Yao et al. \cite{pixel_unet_pansharpening} or pixel-based improved classification of the land cover maps like in Stoian et al. \cite{lclunet}. It has also been tested for fire smoke detection on forests from satellite images \cite{fireunet}.

In the field of urban development, sustainable urban planning and heat stress mitigation examples where this architecture has been used can also be found. For example, McGlinchy et al. \cite{imperviounessUnet}  make use of a U-Net to improve the classification of impervious surfaces in cities. Thanks to better pixel-by-pixel results from fully connected convolutional networks together with satellite imagery, they are able to more accurately map the imperviousness of materials. Along the same lines, Awad et al. \cite{urbanforestUnet} also resort to this technology for classification, in this case, of different forest types in urban and peri-urban areas. In doing so, they merge two machine learning (ML) techniques (artificial neural networks (ANNs) unsupervised self-organizing maps and U-Net) and combine them to create a method referred to as self-organizing-deep-learning. Coupled with satellite imagery, this modeling approach improves the detection of green areas in complex environments. Wang et al. \cite{Wang_2022} go one step further by applying a U-Net-like Transformer for semantic segmentation of different urban scenery, achieving a higher accuracy compared to several state-of-the-art models, both CNN-based and Transformer-based. The model can discriminate between different elements such as low vegetation, cars, buildings and impervious surfaces, among others. 

A particularly relevant contribution to the heat stress mitigation is the work by Su et al. \cite{su2023image}, where they obtain an estimation for the air temperature along all China, training a U-Net with satellite data of Land Surface Temperature (LST). Their modeling approach is proven to faithfully estimate the daily mean, maximum and minimum temperatures, with results validated over data collected by 585 meteorological stations.

\subsection{Shaping of Ground-level Air Temperature}\label{Ta review}

In the context of this paper, temperature shaping refers to the process of estimating air temperature values by learning the underlying spatial and temporal patterns from complex input data. The models herein presented are designed to recognize patterns in the data, such as spatial correlations between different regions or temporal changes over time (like diurnal cycles or seasonal variations). Through this process, the model  ``shapes'' its output to match the real-world temperature dynamics. Achieving a correct T\textsubscript{a} shaping is an important issue for the mitigation of thermal stress in cities. In this subsection we will dig into the different methods that treat the shaping of T\textsubscript{a} in different ways. 

Forecasting and estimating air temperature at ground level is very useful when making decisions to solve urban sustainability problems. For T\textsubscript{a} to be useful for decision makers, the spatial resolution needs to be high enough to be able to differentiate temperature patterns within the city or street itself. However, the most prevalent models used historically for temperature prediction and estimation are numerical models describing atmospheric physics, such as the Numerical Weather Prediction models \cite{numericalwp} or the General Circulation Models (GCMs) \cite{gcm}. 
Nevertheless, these models face limitations in achieving high spatial resolution due to computational constraints and the computational processes are time-consuming, making real-time response not possible. GCMs, for example, typically operate at resolutions ranging from 10 kilometers to several hundred kilometers \cite{flato2014evaluation}. In contrast, the Advanced Regional Prediction System \cite{arps} offers resolutions ranging from approximately 10 kilometers (0.11° × 0.11° grid) to nearly 1 kilometer \cite{paris,resolution}. 

To represent the influence of urban areas in the atmospheric patterns, several numerical schemes of different levels of complexity have been developed in the last years \cite{joshua}. Among them, the model widely known as Urban Climate (UrbClim) is able to resolve urban physics at fine-grained spatial resolution (100 m) and for long time series (e.g. decades), while keeping acceptable accuracy \cite{Urbclim}. An interesting feature of UrbClim is that it uses forcing by a large-scale host model, which reduces dependence of local reference data and allows urban climate projections. The UrbClim model is equipped to simulate various climate variables, including air temperature, humidity and wind speed, on an hourly basis at a spatial resolution of 100 meters. To achieve this, the model relies on extensive meteorological and terrain data as input. This adaptability has led to its integration into numerous studies and projects, such as the assessment of UHI effects in cities like Brussels \cite{LAUWAET20161}, Johannesburg  \cite{SOUVERIJNS2022101331}, and Lisbon  \cite{REIS2022101168} or the computation of heat stress indexes in Seville \cite{sevilla2023variability} and Barcelona \cite{barcelona023mitigation}. It has also acted as a data-source of meteorological variables, mainly air temperature, such as in Fernandez et al. \cite{murcia2023efectos}, where they use the air temperature estimation from UrbClim to analyze the impact of soil sealing on the increase of temperature in urban areas. The same occurs in Sharifi et al. \cite{sharifi2023quantification}, where they use the UrbClim's air temperatures to quantify the future energy impact of global warming on a residential building in Belgium and propose solutions to reduce energy demand. 

Given the versatility of a model such as UrbClim, which provides an accurate estimate of air temperature at ground level at a very fine granularity, similar models that reduce computational time can be a step towards a better efficiency and adoptability of this computationally demanding numerical model. This is why a new field of research has emerged in the last decade, exploring the applicability of AI (especifically, ML models)for the estimation and prediction of T\textsubscript{a} and UHI.

Focusing first on ML models that do not involve deep learning (DL), several works can be found in literature. To begin with, Vulova et al. \cite{summernightberlin} apply different ML algorithms (Random Forest, Stochastic Gradient Boosting and Model Averaged Neural Network) to predict the nocturnal air temperature of Berlin at a spatial resolution of 30 meters. Remote sensing and geodata were used as the training datasets. In the same vein, Varentsov et al. \cite{moscuUHI} also compared different ML algorithms and their accuracy predicting several UHI magnitude series, such as diurnal, synoptic-scale and seasonal variations. Boosting models, in particular CatBoost regression, and Support Vector machines for regression, achieve the most satisfactory results. In their results they also reflect quantitatively the large contribution of urbanization factors to the UHI. Garcia Furuya et al. \cite{suhiSocioeconomic} also took advantage of  ML models to predict LST and surface UHI in Presidente Prudente city (Brasil). By analyzing environmental and socioeconomic variables, they demonstrate the effectiveness of Decision Tree and Random Forest algorithms in optimizing surface UHI characterization, among the several models tested. 

As mentioned above, DL models excel in the area of UHI and T\textsubscript{a}  prediction and estimation. The use of this family of neural computation models is driven by the fact that they can find hidden relationships between air temperature and other meteorological or spatial variables. DL models have been used for several purposes in this field of research, assisting urban planners in decision making processes related to heat stress and UHI mitigation. For instance, estimating and predicting T\textsubscript{a}  can aid in identifying heat-prone areas, optimizing green spaces, and implementing effective cooling strategies. Examples of these applications include finding the relation between green roofs in urban areas and the lowering of the T\textsubscript{a} \cite{greenroof2,greenroof}, determining  the thermal comfort and predicting the T\textsubscript{a}  of green areas in cities \cite{parque1,parque3} and estimating indoor air temperatures and indoor thermal comfort and proving its correlation to UHI \cite{indoor}. In addition, examples can also be found for the explicit calculation and further characterisation of the UHI. As an example of this, the paper of Woo Oh et al. \cite{seoul}, uses feed-forward DNN architecture to gauge a new metric for quantifying the cumulative effects of UHI, called UHI-Hours. They also build a temporal-UHI and a spatial-UHI models, feeding the neural network with LST data, physical and environmental variables. Similarly, Gobakis et al. \cite{predictionUHImodel2011} fed different neural networks models with data on time, ambient temperature and global solar radiation for predicting UHI in Athens, Greece. They accurately predicted air temperature and UHI intensity for at least a 24-hour forecasting horizon using a limited set of data. Assaf et al. \cite{bayesianUHI}, as in the previous example, also worked with ANNs to estimate or predict UHI intensity at census track level scale. However, in this case, they developed a knowledge-based white-box Bayesian neural network model that predicts UHI severity based on demographic, meteorological, and land use factors. Apart from predicting UHI intensity, they also reflected the most important variables affecting the UHI severity. Although UHI is generally treated from a comprehensive urban perspective, there are also examples where more specific cases of UHI are attempted to be predicted. This is the case of Liu et al. \cite{E-uhi_and_industry}, where the focus is on industrial areas. The study analyzes the impact of various factory characteristics on LST and thermal environment. Datasets from land details and field surveys were used to retrieve LST, identify factory structures and classify industrial land. With the use of neural networks, they showcased that the different type of factories, as well as their shape and size, contribute disparately to the surrounding thermal stress. 

The above-mentioned works, while having satisfactory results, lack a fine-grained spatial resolution. It is essential to recognize that urban planners often require higher spatial resolution when dealing with temperature data. The inadequate spatial representation can impact the practical applicability of these findings. Therefore, ongoing efforts to enhance temperature monitoring and modeling at finer scales are crucial for informed decision-making processes in urban planning. On this way, the architecture proposed by Briegel et al. \cite{briegel2023modelling} can open new venues in this area. They employ image-to-image modeling to derive street-level mean radiation temperature (T\textsubscript{mrt}) with a spatial resolution of 1 meter across various zones within Freiburg, Germany. An encoder-decoder architecture, a U-Net specifically, is trained with data comprising meteorological, spatial and urban datasets. The training data consists of 56 patches of the city of 500 × 500 meters, while validation data consists of point measurements taken within 6 test areas. Their approach yields minimal estimation errors for T\textsubscript{mrt}, demonstrating that the architecture proposed is efficient in computation times and robust generalizing across the city under examination.  

\paragraph{Contribution}Following the idea of image regression as a useful method to estimate T\textsubscript{a} with less computational effort and time, the method proposed in this work is built upon a modified U-Net fed with spatial and meteorological variables and has as milestones to determine whether the use of this type of technology gives a new tool to estimate surface air temperature in a much faster way than classical numerical approaches and more effortlessly, with potentials to be more scalable to other cities than the frameworks reviewed above.

\section{Materials and Methods}\label{sec:framework}

The method proposed in this work aims to estimate the T\textsubscript{a} with a spatial resolution of 100 meters in hourly basis. For this purpose a U-Net architecture is proposed, which correlates the different spatial and temporal datasets with the T\textsubscript{a} computed by UrbClim over the grid on which the problem is defined. In the next subsections, the study area (Subsection \ref{sec:studyarea}), the different datasets in use (Subsection \ref{sec:datasets}), and the modeled architecture (Subsection \ref{sec:proposedmodel}) will be presented. 

\subsection{Study Area}\label{sec:studyarea}

The study area, shown in Figure \ref{fig:patches_stations}, covers the hole metropolitan area of the city of Bilbao (Spain) and part of the province of Biscay with an overall area covered of 25 $\times$ 25 $\textup{km}^2$. The area falls within the Cfb (i.e. temperate, no dry season, warm summer) Köppen-Geiger class \cite{beck2018present}. The city is characterized by moderate temperatures in both summer and winter, due to the influence of the Atlantic Ocean. This implies a mean temperature of 14-15 \degree C, reaching maximum temperatures of 25-26 \degree C in July and August \cite{ACERO2015245}. The \emph{Nervión} river flows through the middle of the city and the valley, where the air masses are channeled providing the city with an important ventilation path. The breezes developed by both the topography and proximity of the sea are relevant as well.

\subsection{Datasets}\label{sec:datasets}

In this subsection, we provide an overview of the datasets used to train the proposed model, which integrates both meteorological and spatial information to predict ground-level air temperature. This subsection introduces the various datasets employed, focusing on their structure, sources, and roles in the model training process. Four different datasets are presented to comprehensively describe the data pipeline: the Target Training Dataset, which details the UrbClim-simulated temperature data used as the primary target for the model; the Meteorological Dataset, derived from ERA5 reanalysis data and used to provide meteorological inputs for training; the Spatial Dataset, which comprises various geographical and morphological features that describe the urban landscape of the study area; and finally, the Validation Dataset, which contains real air temperature measurements collected from local weather stations and is used to evaluate the accuracy of the model’s estimations.

\subsubsection{Target Training Dataset}

The target training dataset of the model was based on the dataset called ``Climate variables for cities in Europe from 2008 to 2017'' provided by the Copernicus Climate Change Service \cite{copernicus}. This dataset offers air temperature at a fine-grained resolution of 100 meters in an hourly basis from 2008 to 2017 for a set of 100 European cities. The temperature values were obtained through simulations with UrbClim model \cite{Urbclim}. 

\subsubsection{Meteorological Dataset}

Experiments later presented and discussed in this work utilize meteorological data coming from ERA5 Reanalysis \cite{era5}, which is the same input dataset used by UrbClim. The reanalysis data was facilitated by the European Centre for Medium-Range Weather Forecasts through the Copernicus Climate Data Store in NetCDF file format. ERA5 offers global, hourly data spanning from 1940 to the present at a spatial resolution of 0.25° × 0.25°. Data between 1981 and 2017 was collected for the following surface variables: 2m air temperature, precipitation, specific humidity and wind components (\textit{u} and \textit{v} at 10m). Considering the coarse resolution of ERA5, grid cell values for each variable that fall inside the study area where averaged to obtain the corresponding 1981-2017 hourly time series.  

To reduce the climatic variability of the training dataset and focus the target of the model to those days in which the heat stress can be more present, we applied the Local Weather Type (LWT) classification proposed by \cite{lwt}. The idea is that similar synoptic conditions lead to a similar thermal behaviour of the city and, hence, the model can focus on estimating how the T\textsubscript{a} behaves under those particular conditions (i.e. under that particular LWT). To obtain the LWT several daily metrics are obtained from the above mentioned hourly variables for the baseline period 1981-2010: thermal amplitude (T\textsubscript{a,max} - T\textsubscript{a,min}), total precipitation, average specific humidity, and wind speed and direction. Wind direction was further processed to convert it into a categorical variable by classifying the values based on the following criteria: N (from -45º to 45º), E (from 45º to 135º), S (from 135º to 225º), and W (from 225º to 315º). This classification slightly differs from the one applied by \cite{lwt}, because it follows the recommendations of the Basque Meteorological Agency for the identification of wind components \cite{euskalmet}. We obtained 12 different LWT that were representative of the different seasons. Among them, we selected the one that was most representative for summer (June, July, August and September) and had the potential to generate adverse thermal situations. This cluster was identified as a sunny day with a weak (1.72 m/s) eastern wind component, high specific humidity (around 11 g/kg). After searching the summer days that belong to that specific LWT within the period 2008-2017 we obtained a total of 164 days.

\subsubsection{Spatial Dataset}

The spatial datasets consist of different variables that define the morphological characteristics of the study area. These variables match the ones used by UrbClim to simulate the training dataset \cite{ecfmw}: imperviousness and elevation of the terrain and land cover. The imperviousness and the land cover were obtained from the Copernicus Land Monitoring Service portal \cite{clms}, which provides free European and global geographical information. The DTM was retrieved from the \emph{Instituto Geográfico Nacional} of the Spanish Government \cite{centrodescargas} for convenience. This public repository distributes a DTM product at 200 meters, which is similar to the resolution of the Global Multi-resolution Terrain Elevation Data (GMTED) that is originally used by UrbClim. Although the year of the versions do not match (2012 IGN vs. 2010 GMTED), the impact of the differences can be considered negligible.

Among the variables used originally by UrbClim, two were not considered: the Normalized Difference Vegetation Index (NDVI) and the anthropogenic heat flux. Focusing only on summer days, and considering the NDVI’s seasonal behavior, all the training days are expected to have similar values. In the case of the anthropogenic heat flux, its spatial resolution (5km) was considered too coarse to get the finer spatial patterns derived from the distribution of, for example, traffic and buildings. 

\subsubsection{Train and Test Samples}

Before explaining in detail the architecture used to correlate the T\textsubscript{a} with the spatial and meteorological variables, the subsets of training and test data are described. The first step applied to all the datasets described above was the normalization of the variables in the range [-1,1] in order to improve the interpretability of the data, using:
\vspace{2mm}
\begin{equation}
    x_{norm} = 2 \frac{x-\min x}{\max x - \min x}  -1\ .
\end{equation} 

\vspace{2mm}
Once data was normalized, a subset of training and test data was created by dividing the use case area into several patches, as can be seen in Figure \ref{fig:patches_stations}. A total of 64 patches of 3.1 $\times$ 3.1 km were created. To ensure the representativeness of all map areas during model training, 53 patches were distributed for training, highlighted in black in Figure \ref{fig:patches_stations}. From those 53 patches, 5 of them were reserved for model validation, highlighted in yellow in Figure \ref{fig:patches_stations}. The locations for the test patches were chosen in accordance with the locations of the meteorological stations. The seven different stations lie into six different test patches, with the stations of Deusto and Zorrotza falling into the same patch. The test patches are highlighted in purple in Figure \ref{fig:patches_stations}. Hence, from the 24 hours of the 164 days of LWT,  3936 hours are obtained to train the U-Net model. Consequently, each of the train and test tiles will have 3936 examples. 

\subsubsection{Validation Dataset of Real T\textsubscript{a} Measurements}

\begin{figure}[h!]
    \centering
    \includegraphics[width=\columnwidth]{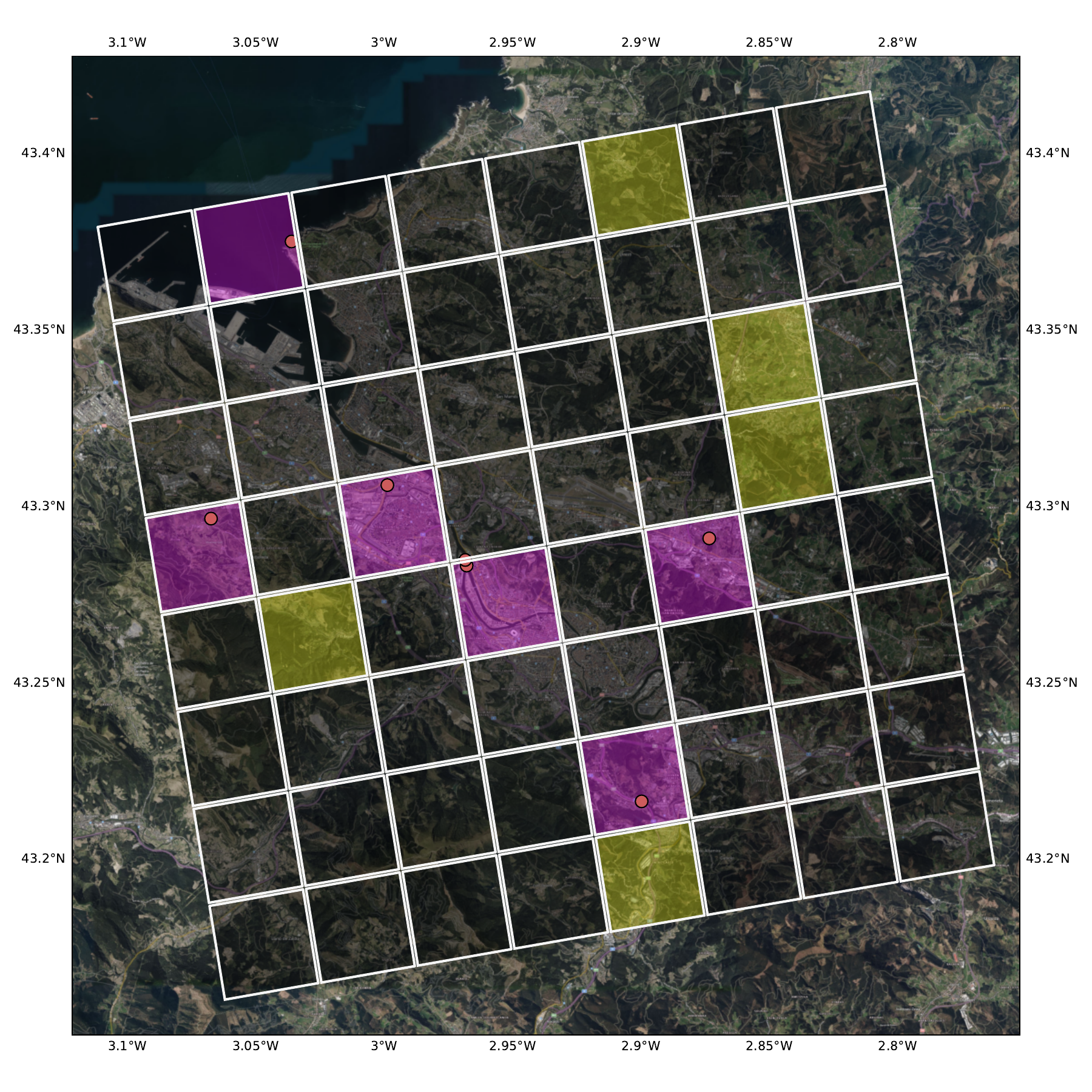}
    \caption{The study area divided in patches. The patches shadowed in yellow correspond to the validation ones while the purple ones to the test patches. The meteorological stations are located in the red dots.}
    \label{fig:patches_stations}
\end{figure}

For validation against real values, we also obtained point measurements from different weather stations that Euskalmet, the Basque meteorological agency, has throughout the study area. T\textsubscript{a} measured values from a total of 7 weather stations were collected for validation in the time range from 2008 to 2017. La Arboleda, located in a mountain; Punta Galea, placed in a lighthouse; Galindo, Deusto and Zorrotza in urban areas near a river; and Arrigorriaga and Derio in urbanized rural areas. Figure \ref{fig:patches_stations} shows the locations of the weather stations marked with a red dot.  

\subsection{Proposed Model: U-Net} \label{sec:proposedmodel}

\begin{figure}[h!]
    \centering
    \includegraphics[width=\columnwidth]{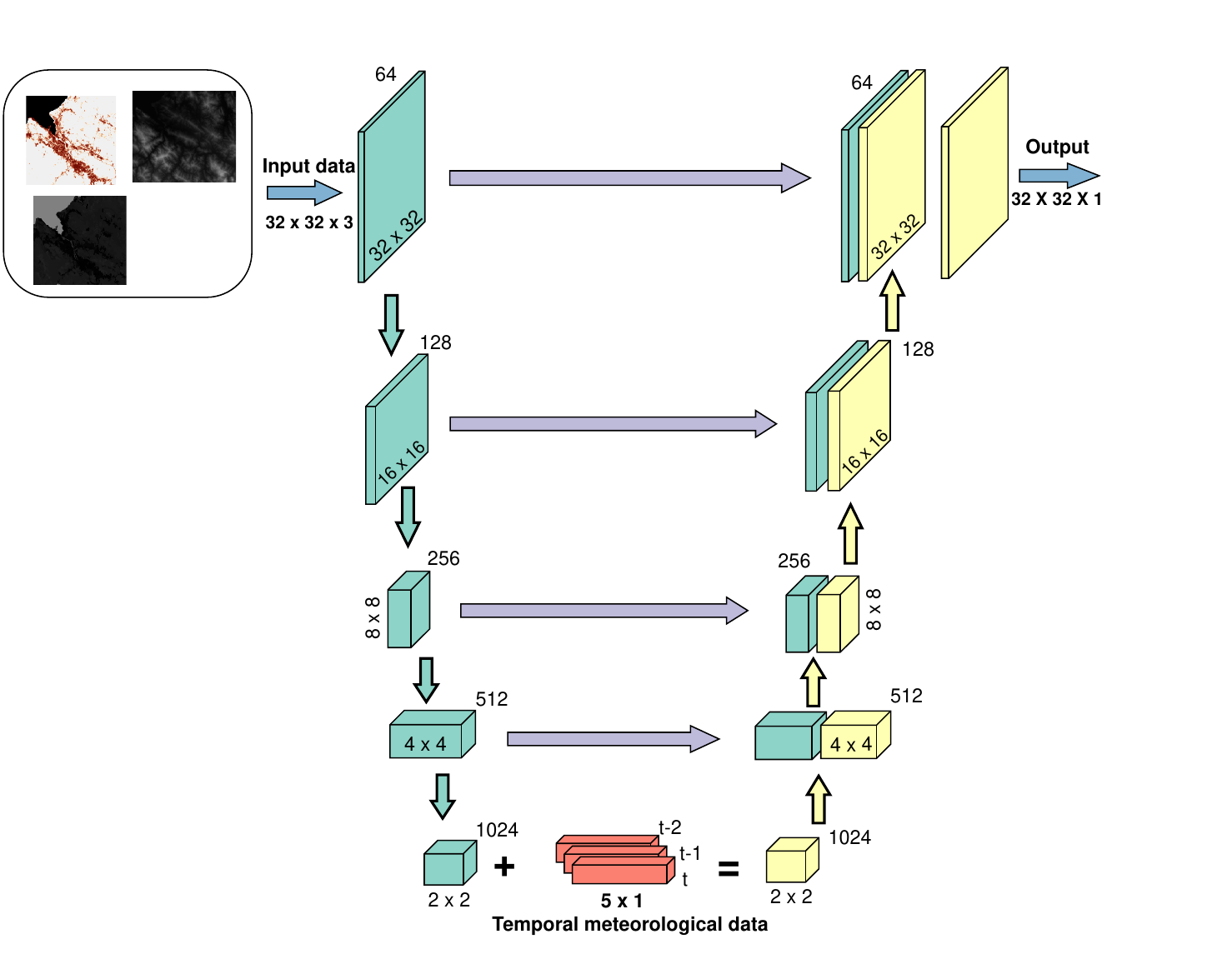}
    \caption{The U-Net architecture proposed for the estimation of T\textsubscript{a}. The sum in the latent space corresponds to a concatenation of two flattened vectors.}
    \label{fig:unet}
\end{figure}

The model used to estimate T\textsubscript{a}, shown in Figure \ref{fig:unet}, is a variation of the well-known U-Net architecture, which consists of an encoder-decoder with skip connections. The input datasets of the encoder consist of the three spatial maps described above. The normalized input tensor representing those variables has the shape of 32 $\times$ 32 $\times$ 3  for each patch of the map.  The U-Net is divided into two parts, the encoder and the decoder. The encoder, composed of CNNs, as well as pooling layers and dropout layers, reduces the dimensionality of the input tensor in aim to find different hidden relationships and extract high level features from it. Once the desired dimension is achieved, the decoder reconstructs the image, applying an inverse process. In this upsampling process, skip connections are added between the encoder and the decoder to help in a high-quality reconstruction of the image. 
For the case described in this work, three downsample blocks are used each consisting of double convolutional layer (with ReLu as the activation function), a pooling layer and a dropout layer. After down-sampling the input data the lowest-dimensional space is reached, also known as latent space. At this stage, due to its low spatial resolution, the meteorological data is added. Both T\textsubscript{a} and the rest of the meteorological variables at time step \textit{t} are conditioned by the meteorological conditions of the previous hours. That is why, to estimate the T\textsubscript{a} of an instant \textit{t}, the meteorological data of the previous 2 hours is taken into account. Hence, the meteorological vectors go from \textit{t-2} to \textit{t}. The meteorological data of each time step is in inserted in a tensor of shape 5 $\times$ 1 to the reconstructed image, resulting in the final output map of the estimated  T\textsubscript{a}. Once the meteorological data is coupled to the spatial data, the upsampling starts. The decoding structure mirrors that of the encoder, comprising 3 up-sample blocks. Each block includes a pair of convolutional transpose layers, a concatenation layer, and a dropout layer and leads to a reconstructed image of shape 32 $\times$ 32 $\times$ 1.

The U-Net contains different parameters that have to be tuned to obtain the optimal performance. For the case exposed here, a learning rate of $10^{-5}$, a batch size of 32, and 30 epochs are chosen. As the \textit{loss function}, Mean Square Error (MSE) is selected: 
\begin{equation}
    MSE = \frac{1}{n} \sum^n_{i=1} (Y_i - \hat{Y}_i)^2
\end{equation}

where $Y_i$ is the observed value and $\hat{Y_i}$ the predicted value. 

\section{Results and Discussion}\label{sec:results}

In the following, we will present and discuss the results obtained for the U-Net model over the scenario defined over the city of Bilbao. The discussed results aim to inform the responses to the three RQs formulated in the introduction. Each of these is addressed individually in the following subsections. For the sake of reproducibility and transparency, source code, result files and plots have been made publicly available in a GitHub repository: \url{https://github.com/InigoD/UHI-UNET}.

\subsection{RQ1: Can a Data-based Model Achieve Accurate T\textsubscript{a} Estimations that are Close to Those Produced by a Numerical Model?}  

In this first RQ the aim is to test whether the estimated values of T\textsubscript{a} are spatially consistent and with values that are close to the ones elicited by the UrbClim numerical model. We recall the reader that, as described in Section \ref{sec:datasets}, the U-Net model is being trained to estimate the T\textsubscript{a} values provided by the Copernicus UrbClim simulations dataset. Hence, a good model will be that reducing the error between the T\textsubscript{a} values obtained from it and those from UrbClim. To assess whether this condition is met by our proposed U-Net architecture, four regression metrics between the two models are computed for each weather station area during the period of study. The metrics chosen are the Pearson Correlation Coefficient (PCC), the Root Mean Square Error (RMSE), the Mean Absolute Error (MAE) and the Mean Absolute Percentage Error (MAPE). The results can be found in Table \ref{tab:metrics}. 

\begin{table}[h!]
    \centering
\resizebox{0.55\columnwidth}{!}{\begin{tabular}{ccccc}
\toprule
 & \multicolumn{4}{c}{UrbClim - U-Net} \\
\cmidrule{2-5}
Station & Pearson & RMSE & MAE & MAPE \\
\midrule
\texttt{arboleda} & 0.98 & 2.71 & 2.13 & 18.83\% \\
\texttt{arrigorriaga} & 0.99 & 2.07 & 1.58 & 14.37\% \\
\texttt{derio} &0.98 & 1.75 & 1.35 & 9.0\%\\
\texttt{deusto} &0.98 & 2.24 & 1.74 & 8.95\% \\
\texttt{galindo} &0.98 & 2.3 & 1.77 & 8.58\% \\
\texttt{puntagalea} &0.95 & 2.51 & 1.95 & 12.47\% \\
\texttt{zorroza} &0.98 & 2.29 & 1.77 & 9.0\% \\
\bottomrule
    \end{tabular}}
    \caption{Regression metrics computed between the two models (UrbClim and U-Net) for each weather station during the period of study.}
    \label{tab:metrics}
\end{table}

As it can be seen in the above table, for all the seven locations, the Pearson correlation is above 0.95, and for almost all of them is near or above 0.98 value. Considering that the definition of the Pearson correlation coefficient is a value between -1 and 1 that quantifies the linear correlation between two datasets, being 0 no correlation and -1 and 1 full correlation depending on the slope, it can be seen that the correlation between both is almost absolute. In other words, when in the reference model, T\textsubscript{a} increases, in the other one also does it in the same or very similar way, and when in the reference model the temperature decreases, so does the proposed model. The location where the Pearson is lower (0.95, \texttt{Punta Galea}) may not be coincidental. Dealing with a cape in the Atlantic Ocean, temperature changes are mostly affected by the sea, making it potentially more challenging for the model to accurately model the patterns of the target variable T\textsubscript{a}. 

We now steer our focus towards the other metrics. In the same table, we observe that both RMSE and MAE are floating around 1.5\degree C to 2.5\degree C of difference. As per its definition, RMSE penalizes more the large errors made by the model. Nonetheless, the highest RMSE is still 2.71\degree C for \texttt{La Arboleda}, an acceptable value. For the MAE, the values are lower than for the RMSE, around 1.7\degree C, being the highest 2.13\degree C also for \texttt{La Arboleda}, and 1.35\degree C the lowest, in \texttt{Derio}. 

Continuing with the quantitative results, Table \ref{tab:regression_metrics} shows a comparison in terms of the same performance metrics between the U-Net and UrbClim models, and the real T\textsubscript{a} measurements collected by the chosen meteorological stations. 

\begin{table}[h!]
    \centering
\resizebox{\columnwidth}{!}{\begin{tabular}{cccccccccc}
\toprule
 & \multicolumn{4}{c}{UrbClim - Real} & & \multicolumn{4}{c}{U-Net - Real} \\
\cmidrule{2-5} \cmidrule{7-10}
Station & Pearson & RMSE & MAE & MAPE & & Pearson & RMSE & MAE & MAPE \\
\midrule
\texttt{arboleda} &0.87 & 7.14 & 5.99 & 33.21\% & & \cellcolor{blue!7}0.88 & \cellcolor{blue!7}5.95 & \cellcolor{blue!7}4.94 & \cellcolor{blue!7}26.68\% \\
\texttt{arrigorriaga} &0.94 & 6.01 & 4.75 & 25.13\% & & \cellcolor{blue!7}0.95 & \cellcolor{blue!7}5.25 & \cellcolor{blue!7}4.28 & \cellcolor{blue!7}23.22\% \\
\texttt{derio} &0.90 & 4.97 & 3.80 & 19.60\% & & \cellcolor{blue!7}0.92 & \cellcolor{blue!7}4.70 & \cellcolor{blue!7}3.63 & \cellcolor{blue!7}18.44\% \\
\texttt{deusto} &0.89 & 6.27 & 4.75 & 22.19\% & & \cellcolor{blue!7}0.91 & \cellcolor{blue!7}5.56 & \cellcolor{blue!7}4.37 & \cellcolor{blue!7}20.98\% \\
\texttt{galindo} &0.89 & 6.77 & 5.10 & 23.26\% & & \cellcolor{blue!7}0.91 & \cellcolor{blue!7}5.84 & \cellcolor{blue!7}4.53 & \cellcolor{blue!7}21.10\% \\
\texttt{puntagalea} &0.87 & \cellcolor{blue!7}3.95 & \cellcolor{blue!7}3.05 & \cellcolor{blue!7}16.40\% & & \cellcolor{blue!7}0.88 & 4.24 & 3.25 & 16.73\% \\
\texttt{zorroza} &0.89 & 5.75 & 4.20 & 20.13\% & & \cellcolor{blue!7}0.91 & \cellcolor{blue!7}4.76 & \cellcolor{blue!7}3.58 & \cellcolor{blue!7}17.33\% \\
\bottomrule
    \end{tabular}}
    \caption{Regression metrics computed for the two models (UrbClim and U-Net) with respect to the real temperature data collected by each weather station during the period of study. Best results for every station and score are highlighted in light blue.}
    \label{tab:regression_metrics}
\end{table}

As can be noticed in the above table, results for the U-Net model are slightly closer to the real measurements collected by all stations, except from \texttt{Punta Galea}. Nonetheless, the results are similar for both models. The UrbClim simulations exhibit spatial artefacts in their T\textsubscript{a} values, particularly in the form of exaggerated temperature gradients. When using the U-Net model for temperature prediction, its convolutional processing inherently smooths these artefacts. Although this smoothing results in a less accurate approximation of the UrbClim output, it has an unintended positive consequence. By blurring the artificial sharp gradients present in the UrbClim data, the U-Net model produces estimates that, despite being a poorer fit to the UrbClim simulations, align more closely with the actual temperature values observed in reality. This suggests that the U-Net’s ability to mitigate spatial artefacts enhances its capacity to generate predictions that are closer to real-world temperature distributions, even though it technically underperforms in replicating UrbClim's output.

The Pearson correlation for the U-Net model is still quite high and encompasses values from 0.88 to 0.95, being most of them around 0.9. The model correlates the T\textsubscript{a} values reasonably well, with MAE for U-Net model with respect to the real T\textsubscript{a} recorded by the weather stations oscillating between 3.25\degree C and 4.94\degree C. It should be noted that the model estimates T\textsubscript{a} with a resolution of 100 meters, while the T\textsubscript{a} values given by the meteorological station are single-point measurements, sometimes collected in urban areas, which may not be as representative as the hole area covered by the estimated pixel of T\textsubscript{a}. This can be regarded as an intrinsic source of error for the evaluation. There is no significant variations in the error between different locations. However, the highest error is located in \texttt{La Arboleda}, which is a mountain of 329 meters above sea level, while the lowest error is in \texttt{Punta Galea}, a meteorological station surrounded by the ocean.

\begin{figure}[h!]
    \centering
    \includegraphics[width=0.49\columnwidth]{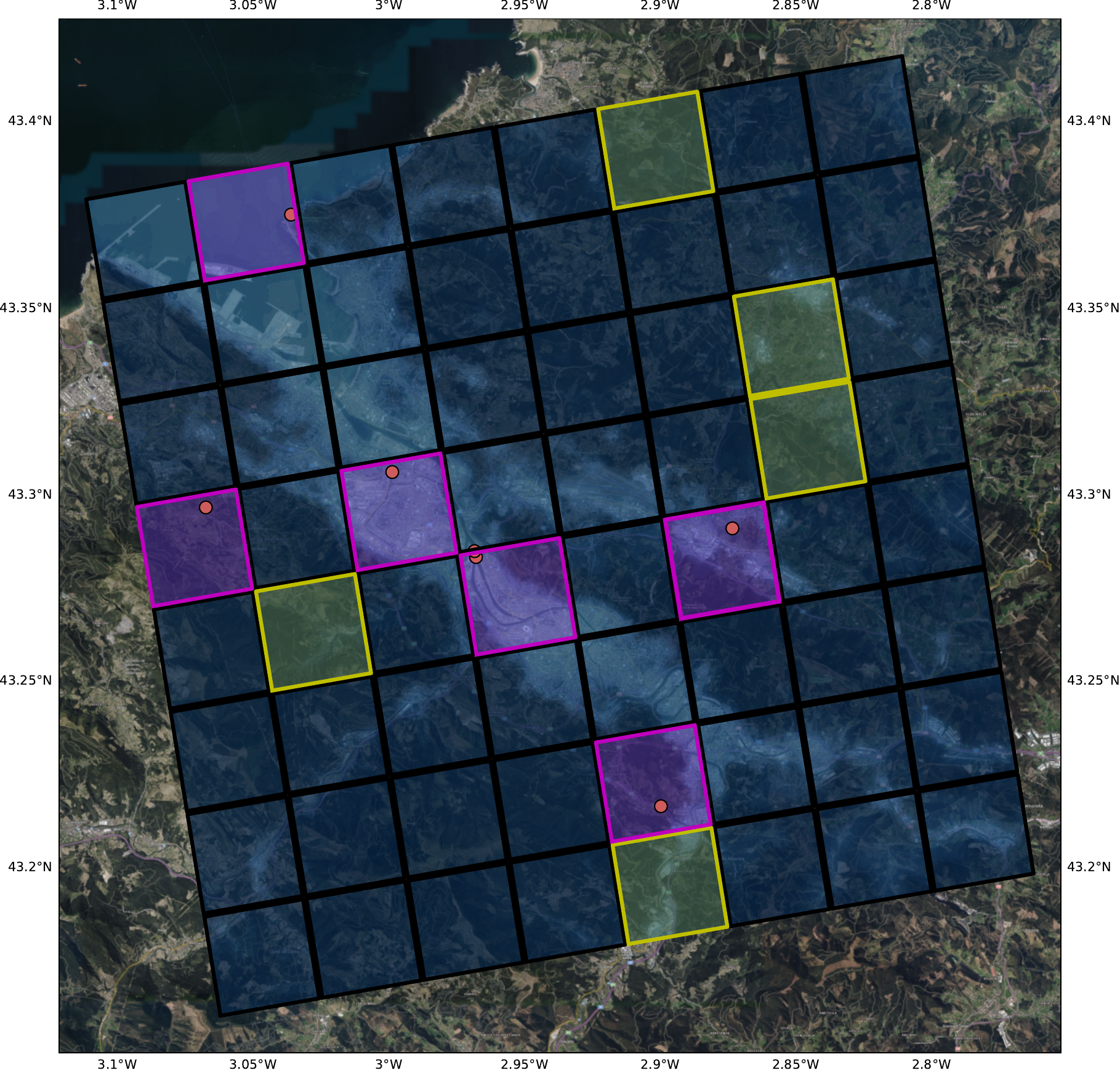}
    \includegraphics[width=0.49\columnwidth]{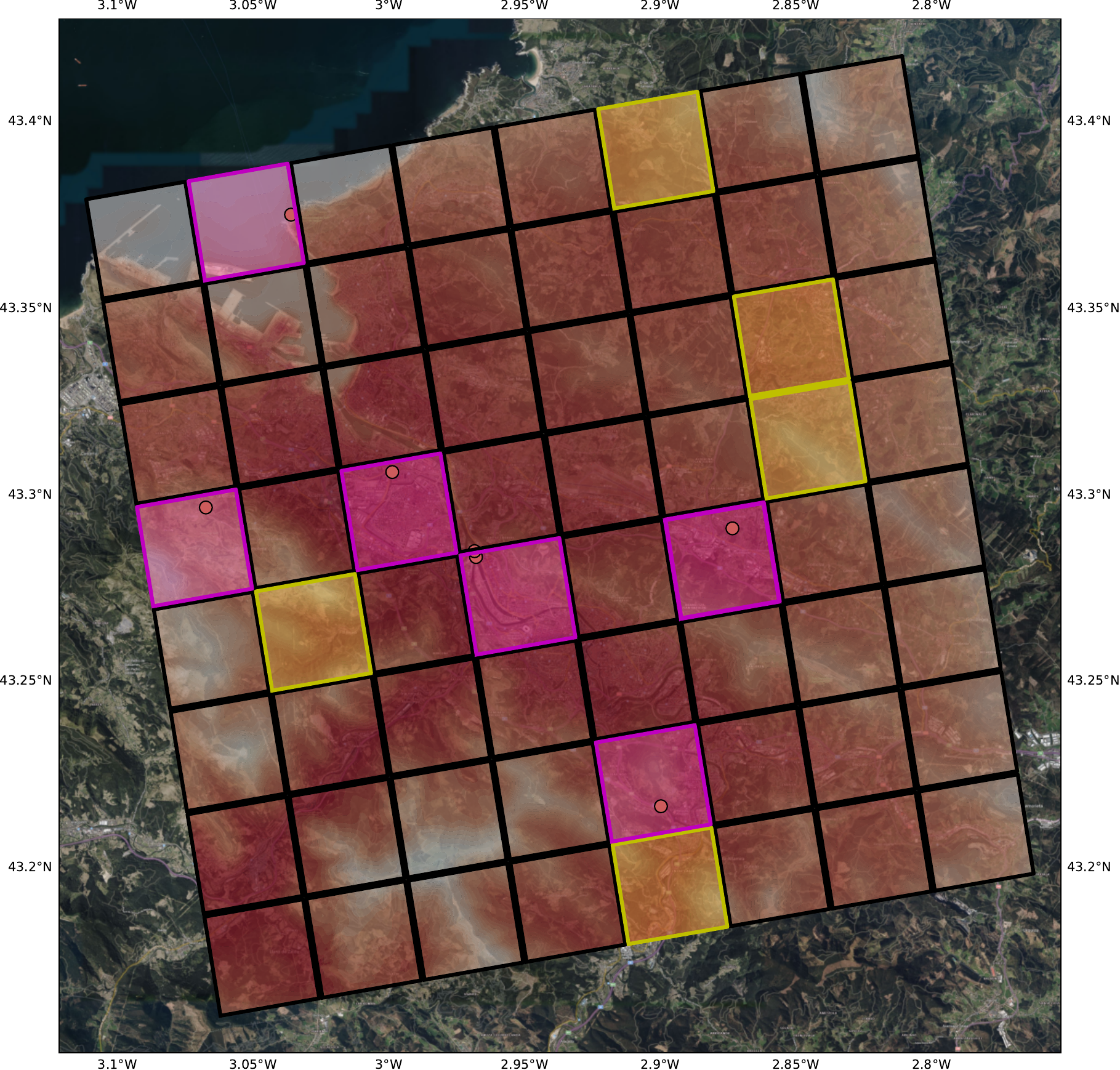}
    \caption{Aggregated spatial distribution U-Net at 05:00 (top) and 14:00 (bottom).}
    \label{fig:spatial_distribution}
\end{figure}

Moving towards qualitative results, Figure \ref{fig:spatial_distribution} shows the aggregated spatial distribution of the predicted T\textsubscript{a} for 05:00 and 14:00 hours. We begin our discussion on the qualitative results with the map corresponding to 5:00, where a spatial consistency of temperatures can be observed in the tiles and their transitions between them. The first thing that can be observed in this plot is the clear influence of the sea as a large body of water that plays the role of temperature reservoir. The areas close to the sea and the estuary maintain milder temperatures (lighter blue). The effect of the sea is clearly and correctly represented.  It can also be observed that the most densely populated areas and where the largest road networks are located are the ones that retain more heat. This is due to the land uses of these areas.  The materials that cover these areas are generally asphalt and concrete, which accumulate more heat during the day. Also, the imperviouness of the land plays an important role. This difference of estimated T\textsubscript{a} as a function of land cover and imperviousness is the expected from the model, since it has been trained taking into account those datasets. Finally, it can be seen that the areas with more vegetation and height such as the mountains are those that reflect lower temperatures (dark blue). Specifically, the Ganekogorta mountain range (1000 meters) is the area where the lowest temperatures are estimated, according to what occurs in practice. Hence, we confirm that the U-Net model learns the relationship between the different spatial datasets fed in the input of the model, such as height, and T\textsubscript{a}.

Following our qualitative analysis, we now focus on the map for the 14:00, which best reflects the thermal contrasts between different zones. The bright colors represent a higher temperature, and as in the early morning map, heat retention in urban areas is much higher than in green areas and mountains. In addition, the geographical idiosincrasy of Bilbao and its surroundings, nestled in a narrow valley, makes thermal differences be very large within short distances. Once again, the Ganekogorta mountain range stands out as the area with the lowest temperatures. In the same way, both the sea and the mountains are very well delimited by the model. At the cape of Punta Galea, in the upper right part, there is a clear change of estimated temperature between the sea area and the land area. 

\begin{figure}[h!]
    \centering
    \includegraphics[width=0.49\columnwidth]{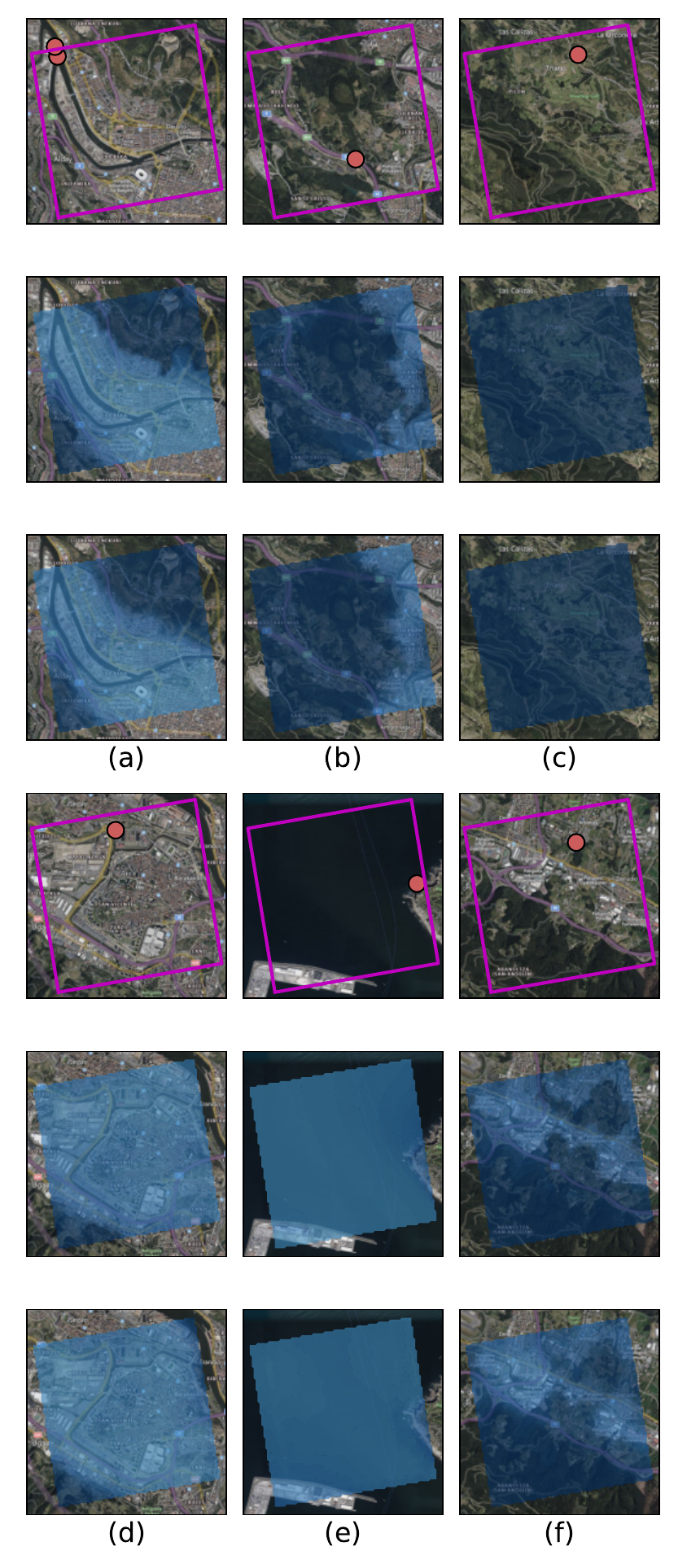}
    \includegraphics[width=0.49\columnwidth]{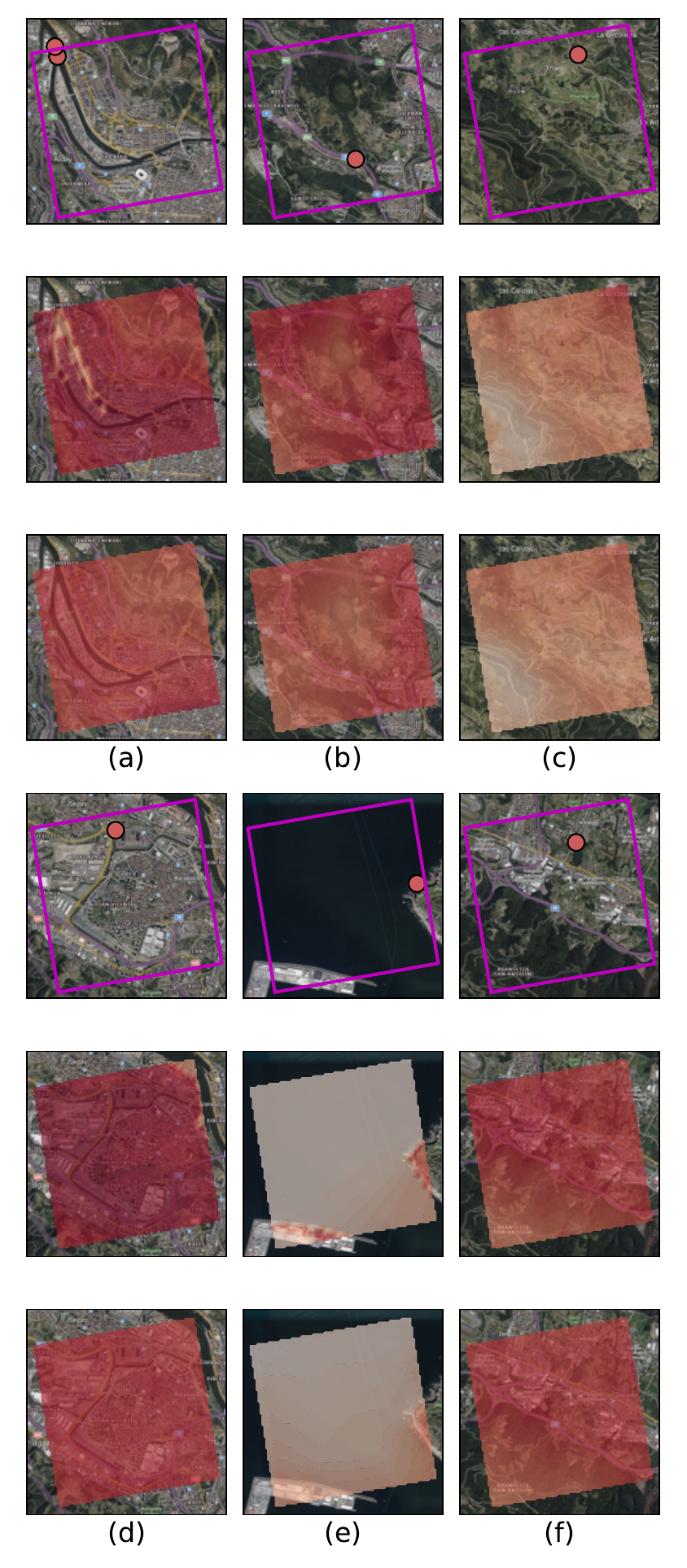}
    \caption{Hourly aggregated T\textsubscript{a} for all the 164 days for the UrbClim model (top) and U-Net model (bottom). In the figure the left image shows the 5:00 aggregation and the right image the 14:00 aggregation.}
    \label{fig:urbclim vs unet}
\end{figure}

Now, we turn the focus towards analyzing in detail the results corresponding to the test tiles, comparing them with the estimations of the UrbClim numerical model. This analysis is done in Figure \ref{fig:urbclim vs unet}, where it is worth noticing the difference in the gradients of T\textsubscript{a}. The first thing that comes to eye in all scenarios is the difference in the gradients of T\textsubscript{a}. The results of the U-Net model show smooth transitions between temperature differences, with more homogeneous values and no abrupt changes. This does not occur in the numerical model. The reason relies on the nature of the convolutional operations that take part in the architecture of the U-Net. The convolution operation, by definition, tends to reduce noise and to smooths gradients spatially.

In the 5:00 maps, the spatial results obtained by the U-Net are very similar to those of the numerical model taken as reference. In the \texttt{Deusto}-\texttt{Zorroza} tile (Figure \ref{fig:urbclim vs unet} (a)), it is able to discern between the temperature of the estuary, its influence in the margins, and the surrounding heights in a very similar way to the numerical model. In \texttt{Derio} (Figure \ref{fig:urbclim vs unet} (f)), the colder zones that appear in the numerical model are also localized by the U-Net model even though they have a complex geography. However, U-Net ensures a smoother spatial gradient of the estimated temperature which, as argued previously, guarantees a more plausible spatial distribution of this variable in the area under study. The same happens in the case of \texttt{Galindo} (Figure \ref{fig:urbclim vs unet} (d)) and \texttt{Arrigorriaga}(Figure \ref{fig:urbclim vs unet} (b)). At \texttt{Punta Galea}(Figure \ref{fig:urbclim vs unet} (e)), the U-Net manages to estimate different temperatures for the small area of land that appears on the right side of the tile. In the 14:00 maps, the same trend continues as in the previous ones. The spatial distribution of the T\textsubscript{a} are very similar in the two models. The U-Net model is able to differentiate small temperature patterns. For example, in the case of \texttt{Arrigorriaga}, in the center of the image can be seen a colder spot which is the representation of the Malmasin mountain that stands out over the whole area. A similar pattern is observed in \texttt{La Arboleda} (Figure \ref{fig:urbclim vs unet}(c)), where the spatial distribution of lower T\textsubscript{a} values aligns with areas of higher altitude and greater vegetation coverage. At \texttt{Punta Galea} the protruding land area is still estimated quite accurately in both models. In the cases of \texttt{Deusto}-\texttt{Zorroza} and \texttt{Galindo}, however, due to the homogenization and smoothing process applied by the U-Net, spatial details are lost that do appear in the numerical model, such as the river part that appears in a lighter orange. Nonetheless, it can not be confirmed without real data measurement, that the sharp gradient in the numerical mode is correct in both cases. 

In conclusion, it is fair to state that the proposed U-Net model is capable of estimating the spatial distribution of T\textsubscript{a}, achieving a plausible and accurate spatial distribution of this variable, and recognizing singular spatial patterns.

\subsection{RQ2: Can a Data-based Model Consistently Estimate Temperature Over Time?}

\begin{figure}[H]
    \centering
    \includegraphics[angle=90, height=0.85\textheight]{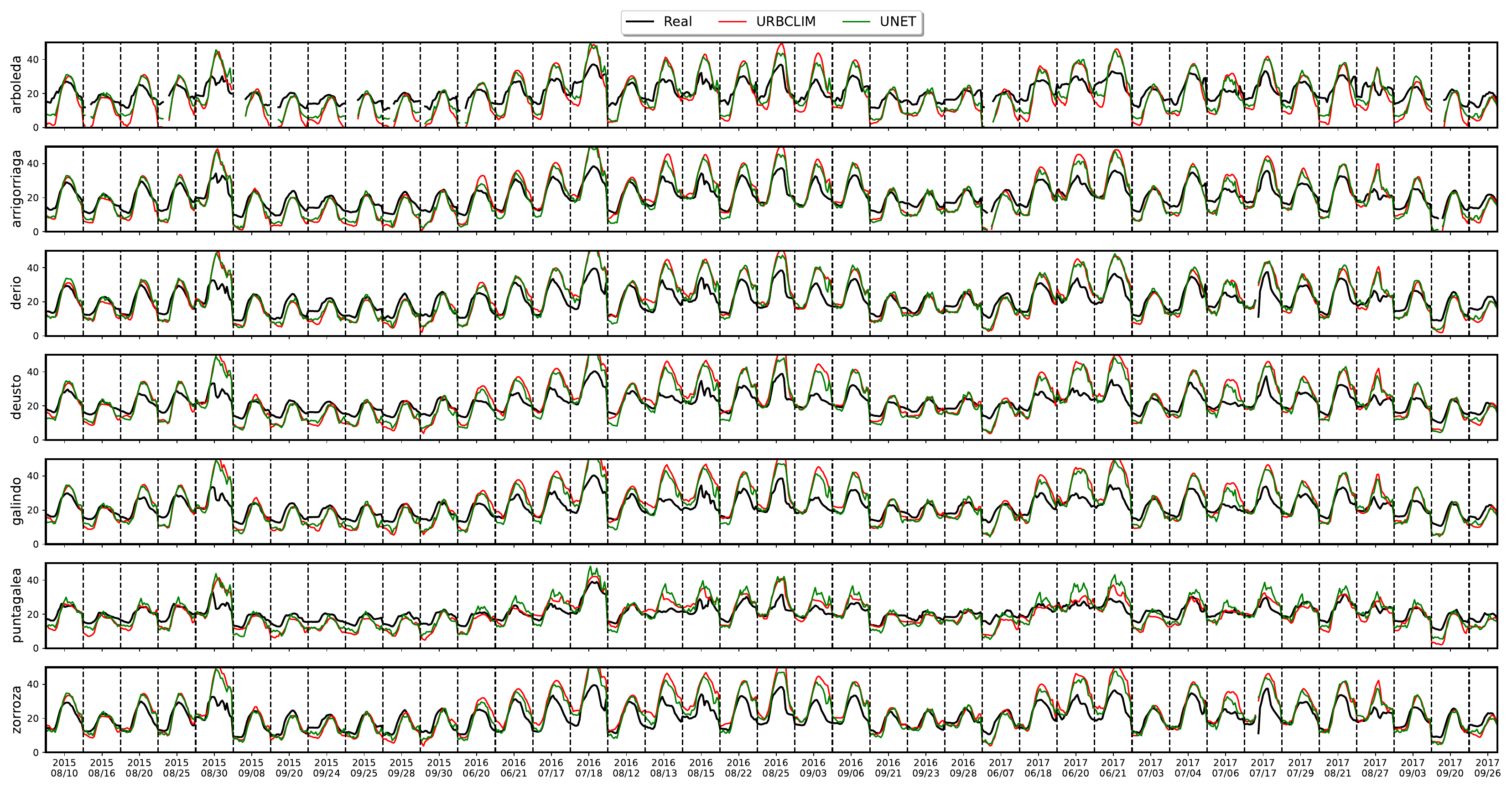}
    \caption{Time series of T\textsubscript{a} for different LWT days, for all the test locations. The values measured by the meteorological stations are represented by the black line. The T\textsubscript{a} values obtained by the UrbClim model are in red, whereas the ones estimated by U-Net model in green.}
    \label{fig:timeseries}
\end{figure}

One of the model's aims is to be able to estimate the T\textsubscript{a} for a resolution of 100 meters and with a temporal frequency of one hour for the 164 days chosen between 2008 and 2017. To answer RQ2, an hourly time series was extracted from the estimated T\textsubscript{a} values for different LWT days. Then, the estimated T\textsubscript{a} values at the same locations as the weather stations were compared both with the weather station data and with the data provided by the UrbClim model. The time series are presented in Figure \ref{fig:timeseries}. 

The trends of the day-night cycles closely resemble those of reality. However, in both models (U-Net and UrbClim) a clear overestimation of the temperature is observed during the day cycle. This overestimation, which in some specific cases such as \texttt{Deusto} or \texttt{Galindo} can reach more than 6\degree C, generally remains at about 3-4\degree C, similar to the errors shown above in Table \ref{tab:metrics}. During the night cycle, estimations vary depending on the weather station and both cases of overestimation and underestimation can be found. 
 
As shown in these plots, the U-Net model effectively reproduces the temperature patterns of the numerical model it was trained to approximate, demonstrating a strong capacity to mirror UrbClim outputs on an hourly basis. A comparison of the respective curves reveals a high degree of similarity, with near-overlapping results observed at certain stations, such as \texttt{Derio}, indicating that the model has accurately learned from the numerical model. Notably, the U-Net model maintains accuracy even when applied to test tiles, highlighting that it is not merely performing temporal interpolation—predicting values based on previous or subsequent time points—but is also handling spatial interpolation effectively.

\begin{figure}[h!]
    \centering
\includegraphics[width=\columnwidth]{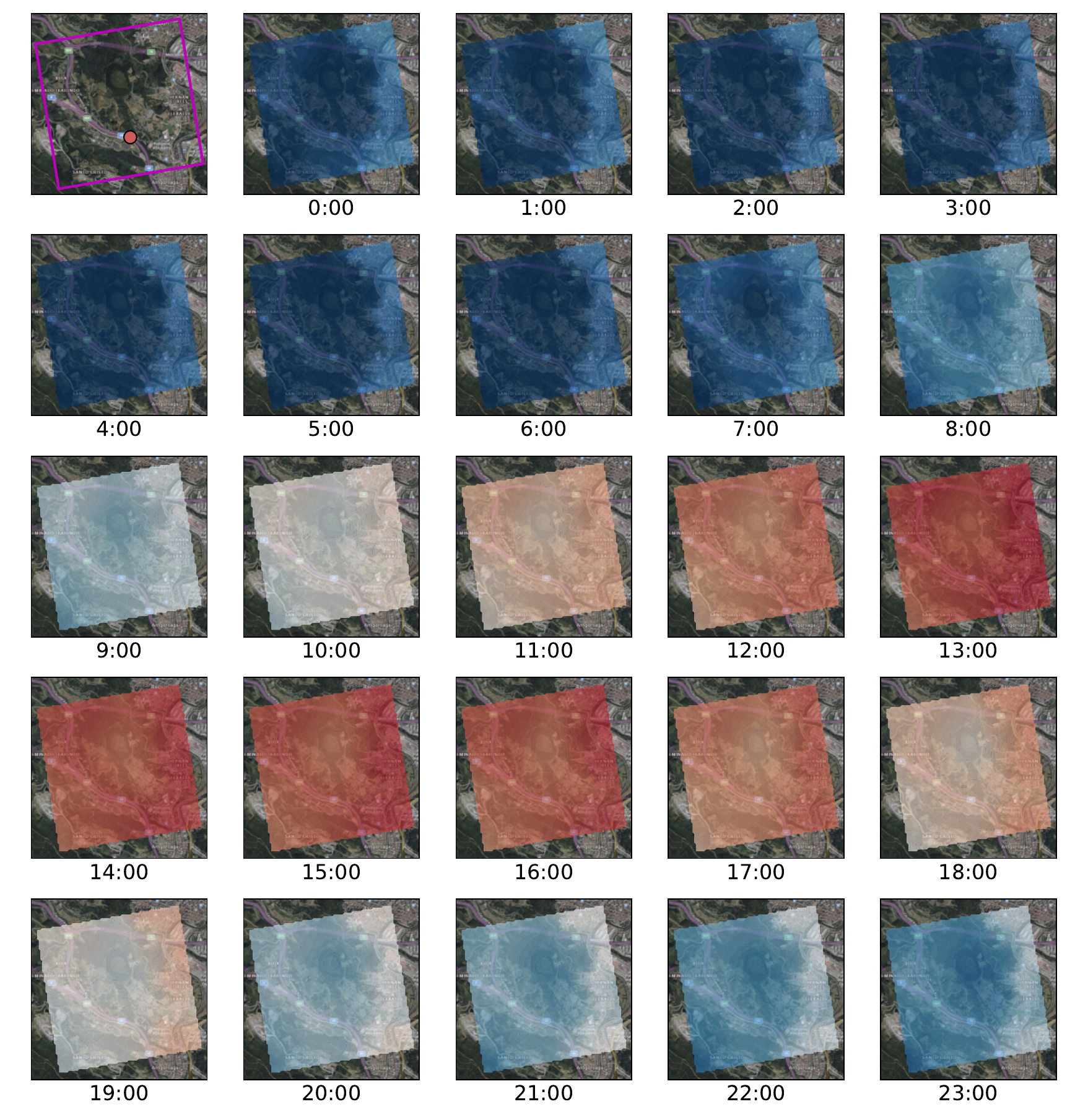}
    \caption{Aggregated temporal progression of the 24 hours for the area of \texttt{Arrigorriaga}.}
    \label{fig:temporal-progression}
\end{figure}

In addition to the temporal series, in Figure \ref{fig:temporal-progression} is shown the aggregated temporal progression of a whole day of one of the test locations, in particular, the one of \texttt{Arrigorriaga}. The temporal progression during the day holds both spatial and temporal consistency. During the whole day, the hottest places are located in the industrial area of the top-right part of the tile, where a steel mill is located. Moreover, in the last hours of the day, after 20:00, the effects of the industrial zone on heat dissipation are clearly visible. The materials used in construction and industry tend to release the heat accumulated during the day to the night, giving rise in many occasions to a nocturnal hotspot. The proposed model correctly reproduces this behavior throughout the day. In addition to this, the fresh spot located in the center of the tile, delineating the Malmasin mountain, is also visible during the whole day. Hence, the model learns to discriminate T\textsubscript{a} between different areas, and moreover the estimation is consistent over time. 

\subsection{RQ3: Can a Data-based Model Estimate Temperature with a Fine-grained Resolution so that Hotspots in Cities can be Identified?} 

\begin{figure}[htpb]
    \centering
    \includegraphics[width=0.8\columnwidth]{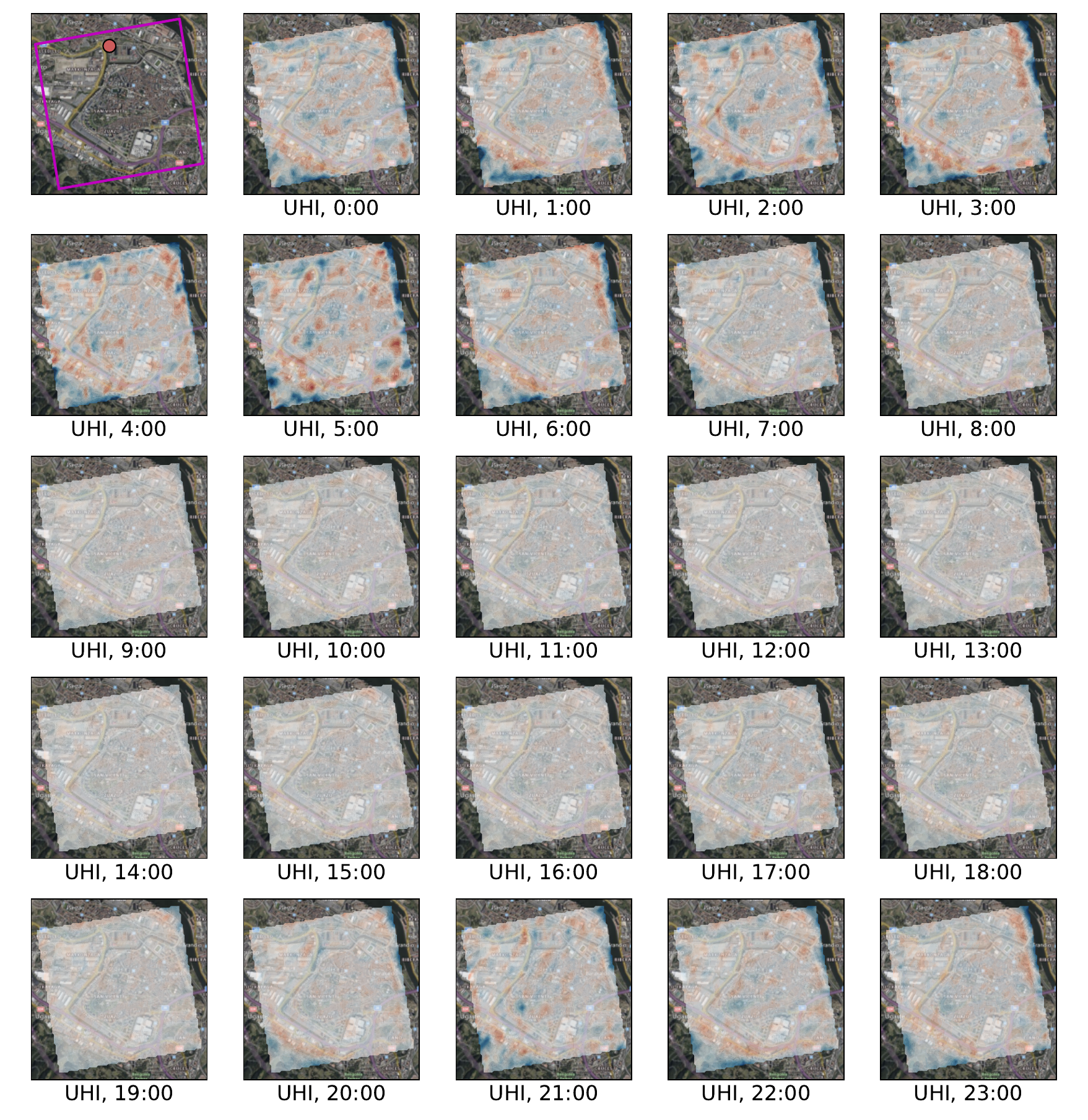}
    \includegraphics[width=\columnwidth]{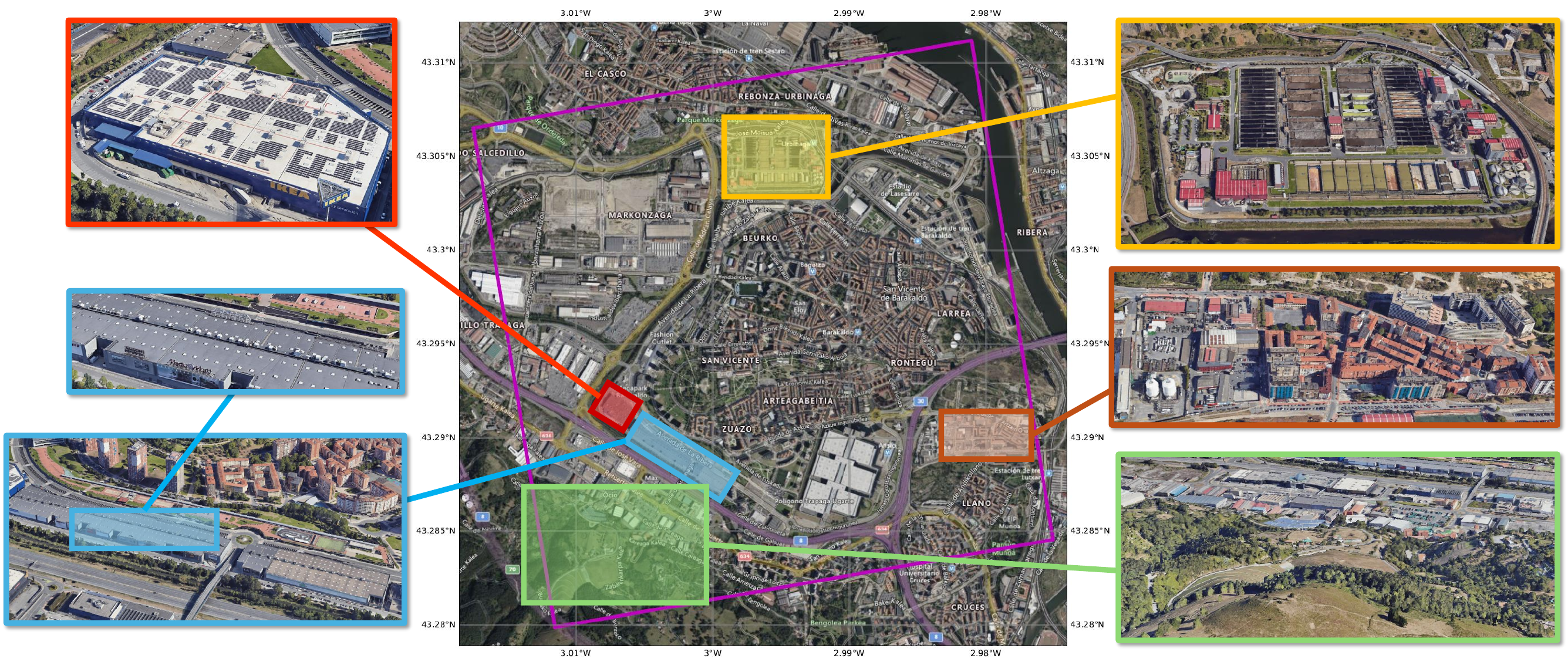}
    \caption{Temporal progression of 24 hours for the relative temperature ($\textup{T}^{rel}_{a}$) of the area of \texttt{Galindo} (top). Location of the hotspots in the area (bottom).}
    \label{fig:galindoUHI}
\end{figure}

\begin{figure}[htpb]
    \centering
    \includegraphics[width = 0.8\columnwidth]{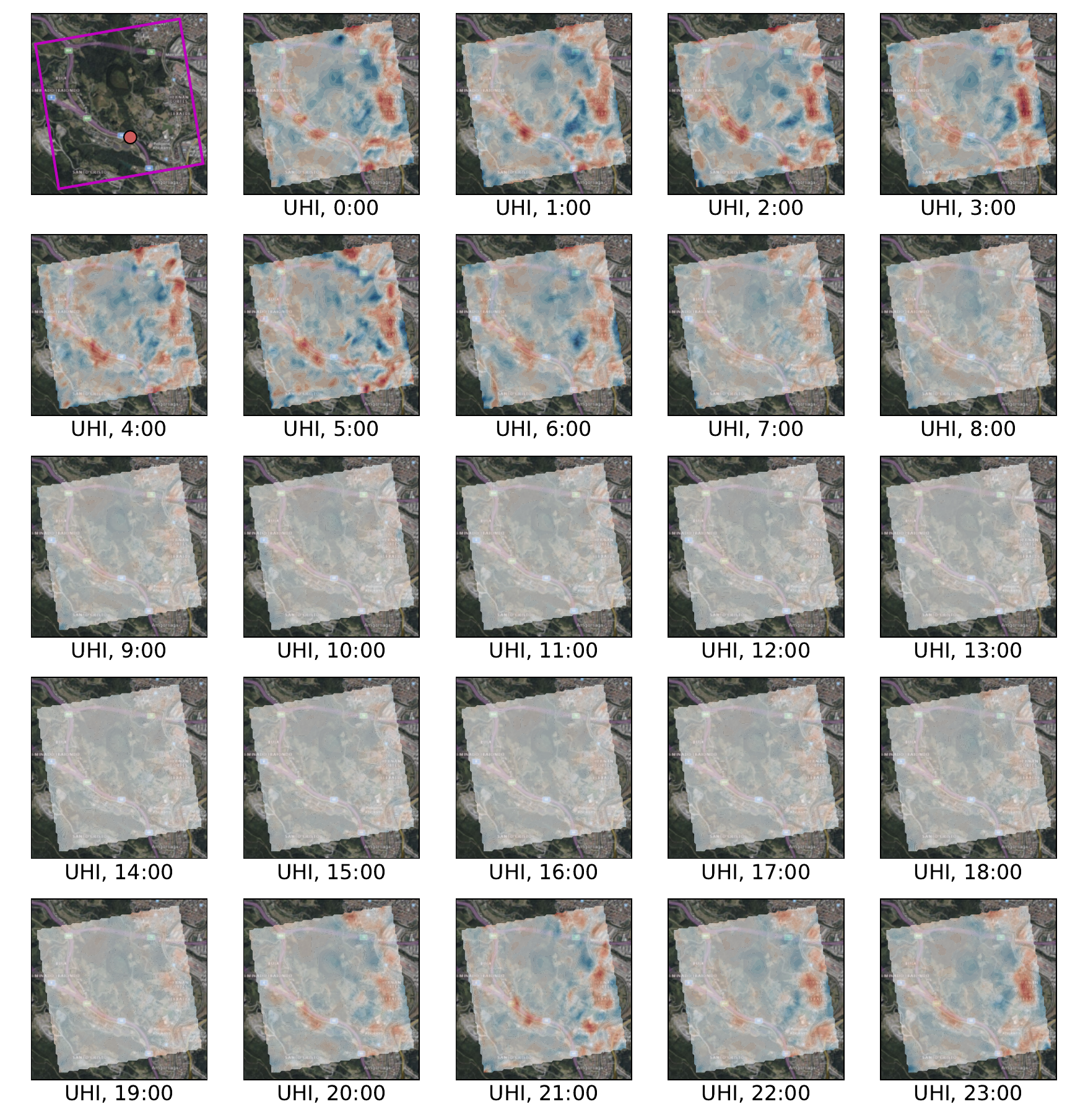}
    \includegraphics[width = \columnwidth]{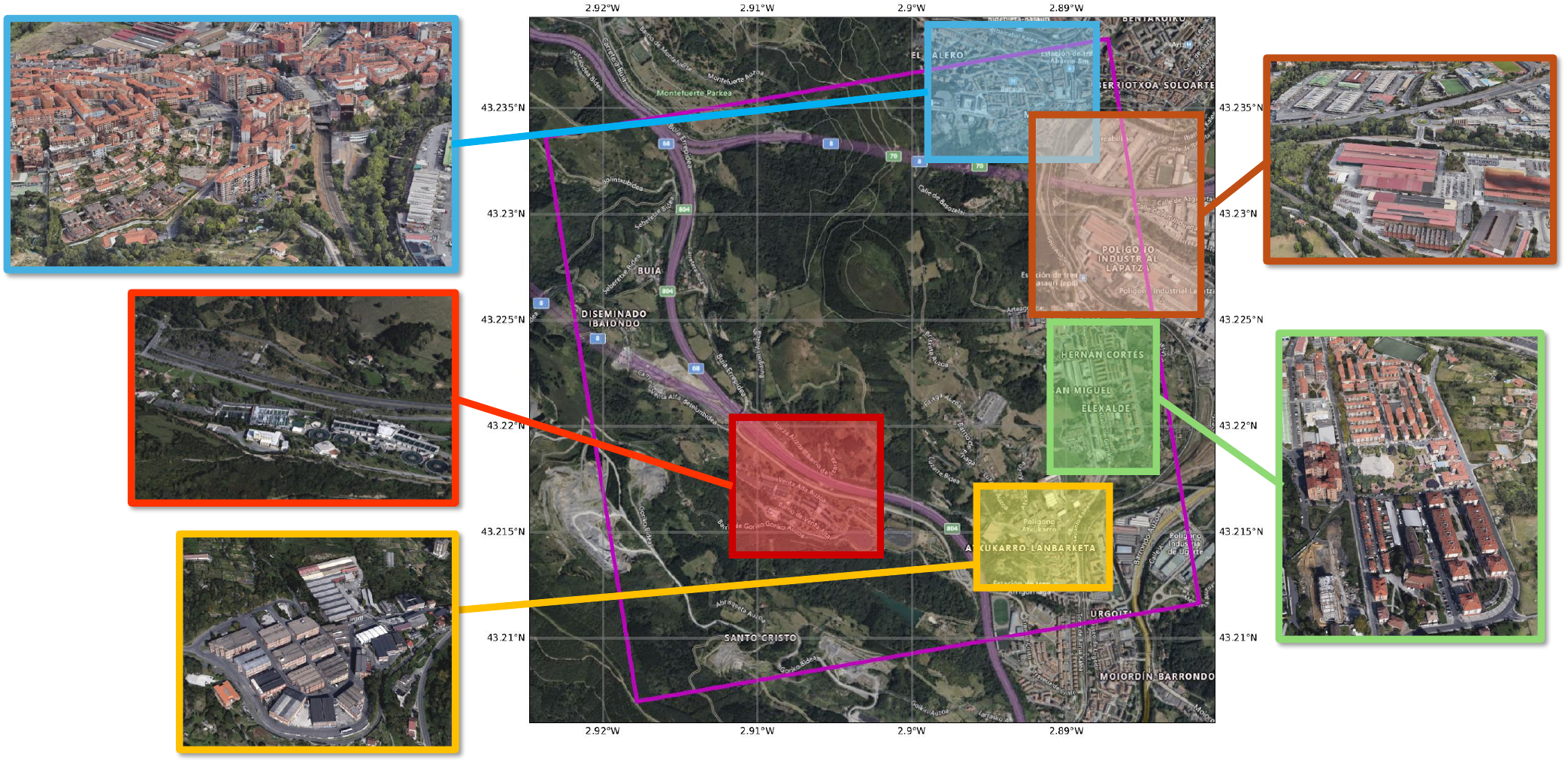}
    \caption{Temporal progression of 24 hours for the relative temperature ($\textup{T}^{rel}_{a}$) of the area of \texttt{Arrigorriaga} (top). Location of the hotspots in the area (bottom).}
    \label{fig:arrigoUHI}
\end{figure}

As it has been explained in the introduction of this paper, a hotspot where urban actions needs to be taken is one that its T\textsubscript{a} stands out steadily over time, in respect to the surrounding areas. Hence, the following methodology is proposed to detect hotspots: for each pixel within a tile, and for each hour and day, the relative difference between its T\textsubscript{a} value and the average T\textsubscript{a} of the surrounding pixels is calculated for each time point,
\begin{equation}
    T^{rel}_{a} = 100 \times \frac{ T_a - \widehat{T}_s}{T_a},
\end{equation}
 
where $\widehat{T}_s$ is the average of the values of the surrounding $3\times 3$ pixels. The final value of the index for each pixel and hour is the median of that value over a range of days under consideration. By virtue of this definition, spatial differences between T\textsubscript{a} values among neighboring pixels are exacerbated. Departing from this mathematical definition of a hotspot, its value for the estimated T\textsubscript{a} values corresponding to two different locations (\texttt{Galindo} and \texttt{Arrigorriaga}) are shown in Figure \ref{fig:galindoUHI} and Figure \ref{fig:arrigoUHI}, respectively. In addition, we depict in the plots a capture of the buildings and activity areas existing in hotspots estimated for these two areas. In both examples, the relative differences between the T\textsubscript{a} of pixels appear at night, according to the theory about the UHI phenomenon, which is said to be a night phenomenon. When the temperature rises, the hottest spots disappear and the differences between temperatures vanish.

In the example of \texttt{Galindo} (Figure \ref{fig:galindoUHI}) the hotspots estimated by the model come from different sources. The first ones, marked in red and blue in Figure \ref{fig:galindoUHI}, are concentrated in a huge commercial zone where several malls and big stores can be found. These type of constructions are prone to suffer from this phenomenon. Apart from the materials they are built with, their rooftops are covered with air conditioners, which are a significant source of heat at night. Another hotspot location is close to a water treatment plant, depicted in yellow in Figure \ref{fig:galindoUHI}. Beside this treatment plant, some industrial buildings can also be found in this location. The last hotspot is the area of Lutxana (depicted in orange). This is a residential zone surrounded by green areas. Furthermore, a liquid air production company is located nearby. Therefore, the heat retained in this area is much higher than in the adjoining areas, resulting in the formation of a hotspot. On the other hand, the blue areas representing colder pixels than the surrounding ones are the ones located closer to the Nervion river and the in the foothills of mount Argalario, also shown in green in Figure \ref{fig:galindoUHI}. This last area is within a higher altitude and with almost no human intervention, being all natural spaces. 

In the case of \texttt{Arrigorriaga} (Figure \ref{fig:arrigoUHI}), hotspots are more noticeable than in the case of the \texttt{Galindo} area, due to the fact that those spots correspond to highly industrialized zones with heavy industry and steel mills. In addition, these areas are surrounded by less urbanized spaces that better mitigate heat, which leads to steeper temperature gradients and hence, to a larger relative differences in the values of T\textsubscript{a} among nearby pixels. Five important areas with hotspots can be detected. Two of them are located in the two residential areas, Elexalde (center-right of the image and highlighted in green in Figure \ref{fig:arrigoUHI} and Basauri (top of the image and highlighted in blue). These areas are densely populated with almost no trees or vegetation and surrounded by industrial parks. As a consequence, these areas do not dissipate heat during the night and are subject to strong heat stress. Other two hotspots (bottom and top-right of the image, shadowed in yellow and orange respectively) are related to heavy industry and industrial parks. The last hotspot (center-left, shadowed in red) is close to the junction of two important highways, which generates a large asphalted surface surrounded by green areas, leading to a high gradient in T\textsubscript{a}. 

\section{Conclusions and Future Work}\label{sec:Conclusions}

In the next decade, cities will inevitably have to implement measures that will lead them to become more sustainable and resilient places. As seen throughout the paper, T\textsubscript{a} plays a pivotal role in this desired urban sustainability. Urban planners face the challenge of implementing actions that produce tangible impacts and benefits, requiring access to as many resources as possible. In this context, having access to fine-grained, hourly T\textsubscript{a} estimations for a city represents a valuable resource for their decision-making processes. Furthermore, if urban planners have access to tools that provide relevant information in a short time, the efficiency and speed of decision-making processes can be significantly improved.
 
In this context, this manuscript has presented a new model for estimating air temperature at ground level, with an accuracy of 100 meters and with a temporal resolution of one hour. This model, based on the U-Net architecture, has proved to be very efficient in estimating the T\textsubscript{a} in a reasonable time, and with a much smaller amount of variables and data than the numerical models. The three RQs posed in the introduction have been answered with satisfactory results. On the one hand, the model has shown that estimates the T\textsubscript{a} with a proper spatial resolution and consistency. The results show an acceptable error and a good correspondence not only at the amplitude level, but also in capturing the intra-days patterns of the variable of interest. The temporal results have evinced that the propsed model excels at estimating the temperature variations occurring in a patch. It can be said, that the proposed model learns correctly to correlate the temperature with the input datasets, both spatial and meteorological. Finally, a new formula for detecting hotspots in cities has been proposed and has been proven to be very useful for detecting zones with higher heat stress within urban areas. 

The promising results reported in this work should not conceal the fact that several aspects of the proposed model have still room for improvement. To begin with, the U-Net model herein proposed has been proven to work under certain conditions. The model has demonstrated its validity just for one scenario but it cannot be ascertained that it is scalable to others cities. Also, it has been trained just for a few specific LWT days of summer. For these days U-Net has exposed a reliable and robust performance when eliciting spatially and temporally coherent estimations of T\textsubscript{a}. However, for a whole year that robustness has yet to be verified. Additionally, the results obtained in comparison to the numerical model may suggest the possibility of entirely replacing the numerical model with a data-based model. However, numerical models cannot be substituted at present, both for the reasons outlined earlier and because the model requires training on an extensive temperature database. At present, the only way to obtain such a large database is through numerical models, since no city has a wide enough sensor coverage to be able to estimate T\textsubscript{a} with a high spatial resolution. However, the model can be useful for a \textit{what if} scenario, since thanks to the computational speed, temperatures can be estimated in a few minutes after applying changes to the input data. For example, it could be used to see the effect on T\textsubscript{a} that a specific urban action would cause, before actually implementing the urban intervention. Another option would be to use the tool to compare different urban modifications and their effects, thus allowing decision makers to choose the most appropriate intervention regarding their impact on the urban heat distribution of the city at hand. 

Besides the aforementioned improvements, the findings discussed here inspire various avenues of research to be explored in the upcoming future. The extension of the temporal range to other seasons should be done in order to see if the model maintains its robustness. Analyzing and broadening the scope of input datasets that are fed into the U-Net model could be another interesting step forward in the research. If the model is to be trained in different seasons apart from summer, NVDI should be considered for example, due to its seasonal variation that could affect its estimations. Another line of research could be to deploy the model in other cities of the same climatic zone to see if it maintains satisfactory results. If the robustness of the model is confirmed in different locations and seasons of the same climatic zone, the logical progression would be to expand the model to other climatic zones, eventually allowing for a total replacement of the numerical model for the zero-shot estimation of the temperature in new urban scenarios.  

\section*{Acknowledgments}

J. Del Ser acknowledges funding support from the Basque Government through the Consolidated Research Group MATHMODE (IT1456-22).

\section*{Declaration of generative AI and AI-assisted technologies in the writing process}

During the preparation of this work the author(s) used ChatGPT in order to refine language, enhance readability, and improve clarity, ensuring the content aligns with the intended standards of communication. After using this tool/service, the author(s) reviewed and edited the content as needed and take(s) full responsibility for the content of the published article.

\bibliographystyle{ieeetr} 
\bibliography{bibliography}

\begin{thebibliography}{10}

\bibitem{un2023report}
U.~N.~D. of~Economic and S.~Affairs, {\em The Sustainable Development Goals
  Report 2023: Special Edition}.
\newblock United Nations, 2023.

\bibitem{sustainablecity}
M.~Jenks and C.~Jones, ``Dimensions of the sustainable city,'' 2009.

\bibitem{wu1996rural}
H.~X. Wu and L.~Zhou, ``Rural-to-urban migration in china,'' {\em Asian-Pacific
  Economic Literature}, vol.~10, no.~2, pp.~54--67, 1996.

\bibitem{lall2006rural}
S.~V. Lall and H.~Selod, {\em Rural-urban migration in developing countries: A
  survey of theoretical predictions and empirical findings}, vol.~3915.
\newblock World Bank Publications, 2006.

\bibitem{un2018}
U.~N.~D. of~Economic and S.~Affairs, {\em World Urbanization Prospects: The
  2018 Revision}.
\newblock United Nations, 2018.

\bibitem{LancetReport2020}
N.~Watts {\em et~al.}, ``The 2020 report of the lancet countdown on health and
  climate change: responding to converging crises,'' {\em The Lancet},
  vol.~397, no.~10269, pp.~129--170, 2021.

\bibitem{heatclimate}
A.~M.~R. Nishimwe and S.~Reiter, ``Using artificial intelligence models and
  degree-days method to estimate the heat consumption evolution of a building
  stock until 2050: A case study in a temperate climate of the northern part of
  europe,'' {\em Cleaner and Responsible Consumption}, vol.~5, p.~100069, 2022.

\bibitem{pollutionsus}
K.~Vasilakopoulou, D.~Kolokotsa, and M.~Santamouris, {\em Cities for Smart
  Environmental and Energy Futures: Urban Heat Island Mitigation Techniques for
  Sustainable Cities}, pp.~215--233.
\newblock 07 2014.

\bibitem{urbanization_pollution_suhi}
M.~Mokarram, F.~Taripanah, and T.~M. Pham, ``Investigating the effect of
  surface urban heat island on the trend of temperature changes,'' {\em
  Advances in Space Research}, vol.~72, no.~8, pp.~3150--3169, 2023.

\bibitem{health1}
M.~Sz{\'e}kely, L.~Carletto, and A.~Garami, ``The pathophysiology of heat
  exposure,'' {\em Temperature}, vol.~2, no.~4, pp.~452--452, 2015.

\bibitem{health2}
Z.~Xu, G.~Fitzgerald, Y.~Guo, B.~Jalaludin, and S.~Tong, ``Impact of heatwave
  on mortality under different heatwave definitions: A systematic review and
  meta-analysis,'' {\em Environment International}, vol.~89-90, pp.~193--203,
  04 2016.

\bibitem{uhi}
H.~H. Kim, ``Urban heat island,'' {\em International Journal of Remote
  Sensing}, vol.~13, no.~12, pp.~2319--2336, 1992.

\bibitem{Urbclim}
K.~De~Ridder, D.~Lauwaet, and B.~Maiheu, ``{UrbClim}--a fast urban boundary
  layer climate model,'' {\em Urban Climate}, vol.~12, pp.~21--48, 2015.

\bibitem{anthropogenicgreenhouse}
H.~Rodhe, ``A comparison of the contribution of various gases to the greenhouse
  effect,'' {\em Science}, vol.~248, no.~4960, pp.~1217--1219, 1990.

\bibitem{oke1982}
T.~R. Oke, ``The energetic basis of the urban heat island,'' {\em Quarterly
  Journal of the Royal Meteorological Society}, vol.~108, no.~455, pp.~1--24,
  1982.

\bibitem{landcoverandUHI}
T.~Adulkongkaew, T.~Satapanajaru, S.~Charoenhirunyingyos, and
  W.~Singhirunnusorn, ``Effect of land cover composition and building
  configuration on land surface temperature in an urban-sprawl city, case study
  in bangkok metropolitan area, thailand,'' {\em Heliyon}, vol.~6, no.~8,
  p.~e04485, 2020.

\bibitem{karimi2023new}
A.~Karimi, P.~Mohammad, A.~Garc{\'\i}a-Mart{\'\i}nez, D.~Moreno-Rangel,
  D.~Gachkar, and S.~Gachkar, ``New developments and future challenges in
  reducing and controlling heat island effect in urban areas,'' {\em
  Environment, Development and Sustainability}, vol.~25, no.~10,
  pp.~10485--10531, 2023.

\bibitem{steeneveld2011quantifying}
G.-J. Steeneveld, S.~Koopmans, B.~Heusinkveld, L.~Van~Hove, and A.~Holtslag,
  ``Quantifying urban heat island effects and human comfort for cities of
  variable size and urban morphology in the netherlands,'' {\em Journal of
  Geophysical Research: Atmospheres}, vol.~116, no.~D20, 2011.

\bibitem{zhou2018impact}
X.~Zhou and H.~Chen, ``Impact of urbanization-related land use land cover
  changes and urban morphology changes on the urban heat island phenomenon,''
  {\em Science of the Total Environment}, vol.~635, pp.~1467--1476, 2018.

\bibitem{encodecosummary}
H.~Zhang, J.~Xu, and J.~Wang, ``Pretraining-based natural language generation
  for text summarization,'' {\em arXiv preprint arXiv:1902.09243}, 2019.

\bibitem{encoderdecosummary2}
L.~Lebanoff, K.~Song, and F.~Liu, ``Adapting the neural encoder-decoder
  framework from single to multi-document summarization,'' {\em arXiv preprint
  arXiv:1808.06218}, 2018.

\bibitem{encodersummary}
Y.~Liu and M.~Lapata, ``Text summarization with pretrained encoders,'' {\em
  arXiv preprint arXiv:1908.08345}, 2019.

\bibitem{machinetranslation1}
K.~Cho, B.~Van~Merri{\"e}nboer, C.~Gulcehre, D.~Bahdanau, F.~Bougares,
  H.~Schwenk, and Y.~Bengio, ``Learning phrase representations using rnn
  encoder-decoder for statistical machine translation,'' {\em arXiv preprint
  arXiv:1406.1078}, 2014.

\bibitem{machinetranslation2}
K.~Cho, B.~Van~Merri{\"e}nboer, D.~Bahdanau, and Y.~Bengio, ``On the properties
  of neural machine translation: Encoder-decoder approaches,'' {\em arXiv
  preprint arXiv:1409.1259}, 2014.

\bibitem{fcn}
J.~Long, E.~Shelhamer, and T.~Darrell, ``Fully convolutional networks for
  semantic segmentation,'' in {\em 2015 IEEE Conference on Computer Vision and
  Pattern Recognition (CVPR)}, pp.~3431--3440, 2015.

\bibitem{segnet}
V.~Badrinarayanan, A.~Kendall, and R.~Cipolla, ``{SegNet}: A deep convolutional
  encoder-decoder architecture for image segmentation,'' {\em IEEE Transactions
  on Pattern Analysis and Machine Intelligence}, vol.~39, no.~12,
  pp.~2481--2495, 2017.

\bibitem{chen2017deeplab}
L.-C. Chen, G.~Papandreou, I.~Kokkinos, K.~Murphy, and A.~L. Yuille, ``Deeplab:
  Semantic image segmentation with deep convolutional nets, atrous convolution,
  and fully connected crfs,'' 2017.

\bibitem{zhao2017pyramid}
H.~Zhao, J.~Shi, X.~Qi, X.~Wang, and J.~Jia, ``Pyramid scene parsing network,''
  2017.

\bibitem{2015unet}
O.~Ronneberger, P.~Fischer, and T.~Brox, ``{U-Net}: Convolutional networks for
  biomedical image segmentation,'' 2015.

\bibitem{unettumour}
R.~Mehta and T.~Arbel, ``{3D U-Net} for brain tumour segmentation,'' in {\em
  International MICCAI Brainlesion Workshop}, pp.~254--266, Springer, 2018.

\bibitem{tumourunet2}
W.~Chen, B.~Liu, S.~Peng, J.~Sun, and X.~Qiao, ``{S3D-UNet}: Separable {3D
  U-Net} for brain tumor segmentation,'' in {\em Brainlesion: Glioma, Multiple
  Sclerosis, Stroke and Traumatic Brain Injuries} (A.~Crimi, S.~Bakas,
  H.~Kuijf, F.~Keyvan, M.~Reyes, and T.~van Walsum, eds.), (Cham),
  pp.~358--368, Springer International Publishing, 2019.

\bibitem{tumourunet3}
A.~Saha, Y.-D. Zhang, and S.~C. Satapathy, ``Brain tumour segmentation with a
  muti-pathway {ResNet} based {UNet},'' {\em Journal of Grid Computing},
  vol.~19, pp.~1--10, 2021.

\bibitem{medicine2020denseunet}
Y.~Cao, S.~Liu, Y.~Peng, and J.~Li, ``Denseunet: densely connected unet for
  electron microscopy image segmentation,'' {\em IET Image Processing},
  vol.~14, no.~12, pp.~2682--2689, 2020.

\bibitem{medicine2022swinUnet}
H.~Cao, Y.~Wang, J.~Chen, D.~Jiang, X.~Zhang, Q.~Tian, and M.~Wang,
  ``Swin-unet: Unet-like pure transformer for medical image segmentation,'' in
  {\em European conference on computer vision}, pp.~205--218, Springer, 2022.

\bibitem{medicine2022afterunet}
X.~Yan, H.~Tang, S.~Sun, H.~Ma, D.~Kong, and X.~Xie, ``After-unet: Axial fusion
  transformer unet for medical image segmentation,'' in {\em Proceedings of the
  IEEE/CVF winter conference on applications of computer vision},
  pp.~3971--3981, 2022.

\bibitem{unet_pavement}
R.~Augustaukas and A.~Lipnickas, ``Pixel-wise road pavement defects detection
  using {U-Net} deep neural network,'' in {\em 2019 10th IEEE International
  Conference on Intelligent Data Acquisition and Advanced Computing Systems:
  Technology and Applications (IDAACS)}, vol.~1, pp.~468--471, 2019.

\bibitem{unet_fabric}
J.~Jing, Z.~Wang, M.~Rätsch, and H.~Zhang, ``Mobile-{Unet}: An efficient
  convolutional neural network for fabric defect detection,'' {\em Textile
  Research Journal}, vol.~92, no.~1-2, pp.~30--42, 2022.

\bibitem{pixel_unet_pansharpening}
W.~Yao, Z.~Zeng, C.~Lian, and H.~Tang, ``Pixel-wise regression using {U-Net}
  and its application on pansharpening,'' {\em Neurocomputing}, vol.~312,
  pp.~364--371, 2018.

\bibitem{lclunet}
A.~Soni, R.~Koner, and V.~G.~K. Villuri, ``{M-UNET}: Modified {U-NET}
  segmentation framework with satellite imagery,'' in {\em Proceedings of the
  Global AI congress 2019}, pp.~47--59, Springer, 2020.

\bibitem{fireunet}
Z.~Wang, P.~Yang, H.~Liang, C.~Zheng, J.~Yin, Y.~Tian, and W.~Cui, ``Semantic
  segmentation and analysis on sensitive parameters of forest fire smoke using
  smoke-unet and landsat-8 imagery,'' {\em Remote Sensing}, vol.~14, no.~1,
  2022.

\bibitem{imperviounessUnet}
J.~McGlinchy, B.~Johnson, B.~Muller, M.~Joseph, and J.~Diaz, ``Application of
  unet fully convolutional neural network to impervious surface segmentation in
  urban environment from high resolution satellite imagery,'' in {\em IGARSS
  2019 - 2019 IEEE International Geoscience and Remote Sensing Symposium},
  pp.~3915--3918, 2019.

\bibitem{urbanforestUnet}
M.~M. Awad and M.~Lauteri, ``Self-organizing deep learning (so-unet)—a novel
  framework to classify urban and peri-urban forests,'' {\em Sustainability},
  vol.~13, no.~10, p.~5548, 2021.

\bibitem{Wang_2022}
L.~Wang, R.~Li, C.~Zhang, S.~Fang, C.~Duan, X.~Meng, and P.~M. Atkinson,
  ``Unetformer: A unet-like transformer for efficient semantic segmentation of
  remote sensing urban scene imagery,'' {\em ISPRS Journal of Photogrammetry
  and Remote Sensing}, vol.~190, p.~196–214, Aug. 2022.

\bibitem{su2023image}
P.~Su, T.~Abera, Y.~Guan, and P.~Pellikka, ``Image-to-image training for
  spatially seamless air temperature estimation with satellite images and
  station data,'' {\em IEEE Journal of Selected Topics in Applied Earth
  Observations and Remote Sensing}, vol.~16, pp.~3353--3363, 2023.

\bibitem{numericalwp}
A.~C. Lorenc, ``Analysis methods for numerical weather prediction,'' {\em
  Quarterly Journal of the Royal Meteorological Society}, vol.~112, no.~474,
  pp.~1177--1194, 1986.

\bibitem{gcm}
P.~P. Saha, K.~Zeleke, and M.~Hafeez, ``Impacts of land use and climate change
  on streamflow and water balance of two sub-catchments of the {Murrumbidgee
  River} in {South Eastern Australia},'' in {\em Extreme Hydrology and Climate
  Variability}, pp.~175--190, 2019.

\bibitem{flato2014evaluation}
G.~Flato {\em et~al.}, ``Evaluation of climate models,'' in {\em Climate change
  2013: the physical science basis. Contribution of Working Group I to the
  Fifth Assessment Report of the Intergovernmental Panel on Climate Change},
  pp.~741--866, Cambridge University Press, 2014.

\bibitem{arps}
M.~Xue, K.~K. Droegemeier, and V.~Wong, ``The advanced regional prediction
  system ({ARPS})--a multi-scale nonhydrostatic atmospheric simulation and
  prediction model. part i: Model dynamics and verification,'' {\em Meteorology
  and Atmospheric Physics}, vol.~75, pp.~161--193, 2000.

\bibitem{paris}
A.~Sarkar and K.~De~Ridder, ``The urban heat island intensity of {Paris}: a
  case study based on a simple urban surface parametrization,'' {\em
  Boundary-layer meteorology}, vol.~138, pp.~511--520, 2011.

\bibitem{resolution}
P.~Berg, O.~Christensen, K.~Klehmet, G.~Lenderink, J.~Olsson, C.~Teichmann, and
  W.~Yang, ``Precipitation extremes in a euro-cordex 0.11° ensemble at hourly
  resolution,'' {\em Natural Hazards and Earth System Sciences Discussions},
  pp.~1--21, 12 2018.

\bibitem{joshua}
M.~J. Lipson {\em et~al.}, ``Evaluation of 30 urban land surface models in the
  urban-plumber project: Phase 1 results,'' {\em Quarterly Journal of the Royal
  Meteorological Society}, vol.~150, no.~758, pp.~126--169, 2024.

\bibitem{LAUWAET20161}
D.~Lauwaet {\em et~al.}, ``Assessing the current and future urban heat island
  of brussels,'' {\em Urban Climate}, vol.~15, pp.~1--15, 2016.

\bibitem{SOUVERIJNS2022101331}
N.~Souverijns {\em et~al.}, ``Urban heat in {Johannesburg and Ekurhuleni, South
  Africa}: A meter-scale assessment and vulnerability analysis,'' {\em Urban
  Climate}, vol.~46, p.~101331, 2022.

\bibitem{REIS2022101168}
C.~Reis, A.~Lopes, and A.~S. Nouri, ``Assessing urban heat island effects
  through local weather types in {Lisbon's} metropolitan area using big data
  from the {Copernicus} service,'' {\em Urban Climate}, vol.~43, p.~101168,
  2022.

\bibitem{sevilla2023variability}
D.~Hidalgo-Garc{\'\i}a and H.~Rezapouraghdam, ``Variability of heat stress
  using the urbclim climate model in the city of seville (spain): mitigation
  proposal,'' {\em Environmental Monitoring and Assessment}, vol.~195, no.~10,
  p.~1164, 2023.

\bibitem{barcelona023mitigation}
D.~Hidalgo~Garc{\'\i}a and J.~Arco~D{\'\i}az, ``Mitigation and resilience of
  local climatic zones to the effects of extreme heat: Study on the city of
  barcelona (spain),'' {\em Urban Science}, vol.~7, no.~4, p.~102, 2023.

\bibitem{murcia2023efectos}
E.~J.~I. Fernandez, M.~A.~R. D{\'\i}az, and A.~P. Morales, ``Efectos del
  sellado del suelo en la temperatura del aire en la ciudad de murcia,'' in
  {\em Geograf{\'\i}a: cambios, retos y adaptaci{\'o}n: libro de actas. XVIII
  Congreso de la Asociaci{\'o}n Espa{\~n}ola de Geograf{\'\i}a, Logro{\~n}o, 12
  al 14 de septiembre de 2023}, pp.~375--384, Asociaci{\'o}n Espa{\~n}ola de
  Geograf{\'\i}a, 2023.

\bibitem{sharifi2023quantification}
M.~Sharifi, M.~H. Shamsi, Y.~Ma, and D.~Lauwaet, ``Quantification of the impact
  of global warming on summer overheating risk in a residential building in
  urban areas in belgium,'' in {\em Journal of Physics: Conference Series},
  vol.~2600, p.~092016, IOP Publishing, 2023.

\bibitem{summernightberlin}
S.~Vulova, F.~Meier, D.~Fenner, H.~Nouri, and B.~Kleinschmit, ``Summer nights
  in berlin, germany: Modeling air temperature spatially with remote sensing,
  crowdsourced weather data, and machine learning,'' {\em IEEE Journal of
  Selected Topics in Applied Earth Observations and Remote Sensing}, vol.~13,
  pp.~5074--5087, 2020.

\bibitem{moscuUHI}
M.~Varentsov, M.~Krinitskiy, and V.~Stepanenko, ``Machine learning for
  simulation of urban heat island dynamics based on large-scale meteorological
  conditions,'' {\em Climate}, vol.~11, no.~10, 2023.

\bibitem{suhiSocioeconomic}
M.~Furuya, D.~Furuya, L.~Y. Oliveira, P.~Silva, R.~Cicerelli, W.~Gonçalves,
  J.~Junior, L.~Osco, and A.~P. Ramos, ``A machine learning approach for
  mapping surface urban heat island using environmental and socioeconomic
  variables: a case study in a medium-sized brazilian city,'' {\em
  Environmental Earth Sciences}, vol.~82, p.~325, 06 2023.

\bibitem{greenroof2}
D.~Erdemir and T.~Ayata, ``Prediction of temperature decreasing on a green roof
  by using artificial neural network,'' {\em Applied Thermal Engineering},
  vol.~112, pp.~1317--1325, 2017.

\bibitem{greenroof}
A.~Asadi, H.~Arefi, and H.~Fathipoor, ``Simulation of green roofs and their
  potential mitigating effects on the urban heat island using an artificial
  neural network: A case study in austin, texas,'' {\em Advances in Space
  Research}, vol.~66, pp.~1846--1862, 07 2020.

\bibitem{parque1}
S.~Chan and C.~Chau, ``Development of artificial neural network models for
  predicting thermal comfort evaluation in urban parks in summer and winter,''
  {\em Building and Environment}, vol.~164, p.~106364, 08 2019.

\bibitem{parque3}
F.~Tomatis, F.~J. Diez, M.~S. Wilhelm, and L.~M. Navas-Gracia, ``Prediction of
  daily ambient temperature and its hourly estimation using artificial neural
  networks in urban allotment gardens and an urban park in valladolid, castilla
  y leon, spain,'' {\em Agronomy}, vol.~14, no.~1, 2024.

\bibitem{indoor}
A.~Ashtiani, P.~Mirzaei, and F.~Haghighat, ``Indoor thermal condition in urban
  heat island: Comparison of the artificial neural network and regression
  methods prediction,'' {\em Energy and Buildings}, vol.~76, pp.~597--604, 06
  2014.

\bibitem{seoul}
J.~W. Oh, J.~Ngarambe, P.~N. Duhirwe, G.~Y. Yun, and M.~Santamouris, ``Using
  deep-learning to forecast the magnitude and characteristics of urban heat
  island in {Seoul Korea},'' {\em Scientific Reports}, vol.~10, no.~1,
  pp.~1--13, 2020.

\bibitem{predictionUHImodel2011}
K.~Gobakis, D.~Kolokotsa, A.~Synnefa, M.~Saliari, K.~Giannopoulou, and
  M.~Santamouris, ``Development of a model for urban heat island prediction
  using neural network techniques,'' {\em Sustainable Cities and Society},
  vol.~1, no.~2, pp.~104--115, 2011.

\bibitem{bayesianUHI}
G.~Assaf, X.~Hu, and R.~H. Assaad, ``Predicting urban heat island severity on
  the census-tract level using bayesian networks,'' {\em Sustainable Cities and
  Society}, vol.~97, p.~104756, 2023.

\bibitem{E-uhi_and_industry}
S.~Liu, J.~Zhang, J.~Li, Y.~Li, J.~Zhang, and X.~Wu, ``Simulating and
  mitigating extreme urban heat island effects in a factory area based on
  machine learning,'' {\em Building and Environment}, vol.~202, p.~108051,
  2021.

\bibitem{briegel2023modelling}
F.~Briegel, O.~Makansi, T.~Brox, A.~Matzarakis, and A.~Christen, ``Modelling
  long-term thermal comfort conditions in urban environments using a deep
  convolutional encoder-decoder as a computational shortcut,'' {\em Urban
  Climate}, vol.~47, p.~101359, 2023.

\bibitem{beck2018present}
H.~E. Beck, N.~E. Zimmermann, T.~R. McVicar, N.~Vergopolan, A.~Berg, and E.~F.
  Wood, ``Present and future köppen-geiger climate classification maps at 1-km
  resolution,'' {\em Scientific data}, vol.~5, no.~1, pp.~1--12, 2018.

\bibitem{ACERO2015245}
J.~A. Acero and K.~Herranz-Pascual, ``A comparison of thermal comfort
  conditions in four urban spaces by means of measurements and modelling
  techniques,'' {\em Building and Environment}, vol.~93, pp.~245--257, 2015.

\bibitem{copernicus}
H.~Hooyberghs, J.~Berckmans, F.~Lefebre, and K.~{De Ridder}, ``Climate
  variables for cities in europe from 2008 to 2017.,'' {\em Copernicus Climate
  Change Service (C3S) Climate Data Store (CDS)}, 2019.

\bibitem{era5}
H.~Hersbach {\em et~al.}, ``The {ERA5} global reanalysis,'' {\em Quarterly
  Journal of the Royal Meteorological Society}, vol.~146, no.~730,
  pp.~1999--2049, 2020.

\bibitem{lwt}
J.~Hidalgo and R.~Jougla, ``On the use of local weather types classification to
  improve climate understanding: An application on the urban climate of
  toulouse,'' {\em PLOS ONE}, vol.~13, pp.~1--21, 12 2018.

\bibitem{euskalmet}
``Euskalmet.''
  \url{https://www.euskalmet.euskadi.eus/divulgacion/manual-de-estilo/} (Date
  accessed: 13.10.2024).

\bibitem{ecfmw}
``European centre for medium-range weather forecasts.''
  \url{https://confluence.ecmwf.int/display/CKB/Climate+variables+for+cities+in+Europe+from+2008+to+2017+documentation}
  (Date accessed: 13.10.2024).

\bibitem{clms}
{European Environment Agency, DG Joint Research Centre of the European
  Commission}, ``Copernicus land monitoring service,''
  \url{https://land.copernicus.eu/en}.

\bibitem{centrodescargas}
{Organismo Autónomo Centro Nacional de Información Geográfica (CNIG)},
  ``Centro de descargas,''
  \url{https://centrodedescargas.cnig.es/CentroDescargas/} (Date accessed:
  13.10.2021).

\end{thebibliography}
\end{document}